\documentclass{article}

\usepackage{microtype}
\usepackage{graphicx}
\usepackage{subcaption}
\usepackage{booktabs}
\usepackage{enumitem}
\usepackage{float}
\usepackage{placeins}
\usepackage{hyperref}

\usepackage[accepted]{icml2026}
\usepackage{amsmath}
\usepackage{amssymb}
\usepackage{mathtools}
\usepackage{amsthm}
\usepackage[capitalize,noabbrev]{cleveref}

\theoremstyle{plain}

\theoremstyle{definition}

\theoremstyle{remark}

\newcommand{\lgatr}{L-GATr}
\newcommand{\slim}{L-GATr-slim}
\renewcommand{\part}{ParT}
\newcommand{\vanilla}{Vanilla-T}
\newcommand{\lloca}{LLoCa-T}
\newcommand{\hmv}{h_{\mathrm{mv}}}
\newcommand{\hs}{h_{\mathrm{s}}}
\newcommand{\dauc}{\Delta\mathrm{AUC}}
\newcommand{\bootci}[2]{#1\,[#2]}
\newcommand{\bootval}[2]{#1}
\newcommand{\asymci}[3]{#1\,\genfrac{}{}{0pt}{1}{+#2}{-#3}}

\icmltitlerunning{What Do Lorentz-Equivariant Jet Taggers Learn?}

\begin{document}

\twocolumn[
  \icmltitle{What Do Lorentz-Equivariant Jet Taggers Learn?}

  \icmlsetsymbol{equal}{*}

  \begin{icmlauthorlist}
    \icmlauthor{Jay Agarwal}{bits}
    \icmlauthor{Siddharth Khare}{bits}
    \icmlauthor{Dhruv Kumar}{bits}
  \end{icmlauthorlist}

  \icmlaffiliation{bits}{BITS Pilani, Pilani, India}
  \icmlcorrespondingauthor{Jay Agarwal}{f20221305@pilani.bits-pilani.ac.in}

  \icmlkeywords{jet tagging, geometric algebra, Lorentz equivariance, interpretability,
    linear probes, grade decomposition}

  \vskip 0.3in
]

\printAffiliationsAndNotice{}

\begin{abstract}
We study what Lorentz-equivariant jet taggers learn internally, using equivariance
tests, linear probes and grade ablations across five models including \lgatr{},
\slim{} and \lloca{}.
Linear probes show that equivariant models suppress frame-dependent pseudorapidity
to zero while encoding jet mass and N-subjettiness strongly.
Grade ablations on \lgatr{} reveal that bivector channels are negligible for
top-quark tagging while vector-like channels are dominant but seed-variable,
consistent with the network exploiting multiple representational pathways.
These results characterize which physical features and algebraic grade
structures carry discriminative information in equivariant taggers and may
inform future development of such models.
\end{abstract}

\section{Introduction}
\label{sec:intro}

Equivariant neural networks are becoming central tools for scientific machine
learning because they build known physical symmetries directly into the model.
In high-energy physics (HEP), Lorentz symmetry is especially important: jets
are observed in a particular detector frame, but the underlying physics is
constrained by relativistic transformations. Recent Lorentz-equivariant
architectures achieve strong performance on jet tagging benchmarks, yet high
accuracy alone does not tell us what physics these models have learned, whether
their symmetry constraints are active in practice, or how their internal
representations differ from ordinary transformers.

This question is timely for trustworthy AI in science. Collider analyses
increasingly rely on neural networks to extract subtle structure from complex
events. If symmetry-aware models are to become reliable scientific instruments,
we need tools that inspect not only their outputs, but also whether their
internal representations align with the structure of the physical problem.

We study the Lorentz-equivariant Geometric Algebra Transformer
(\lgatr{}~\citep{spinner2024lgatr}), which builds on the Geometric Algebra
Transformer framework~\citep{brehmer2023gatr} by representing jet constituents
as multivectors in the geometric algebra of Minkowski spacetime. This
representation decomposes into grades: scalars, vectors, bivectors,
trivectors and pseudoscalars. The grade structure gives \lgatr{} an unusual
advantage for interpretability: unlike a generic hidden vector, each component
has a transformation law and a geometric meaning.

We study \lgatr{} and four comparison models: \slim{},
\lloca{}, \vanilla{} and \part{}, using
equivariance checks, linear probes, grade ablations and
invariant multivector probes on TopTagging. Our main contributions are:

\textbf{(1) Linear probes reveal invariant physical representations.}
Linear probes show that all three equivariant architectures suppress
frame-dependent pseudorapidity to zero while strongly encoding jet
mass and N-subjettiness, a representation-level signature of the imposed
symmetry that non-equivariant baselines do not exhibit.
\lloca{} achieves the highest probe scores of all five models for most
physics targets, suggesting that probe accessibility reflects representational
strategy rather than task utility.

\textbf{(2) Grade ablations reveal multiple usable pathways.}
Ablating grade groups within \lgatr{}'s multivector hidden state shows that
one algebraic grade (bivectors) is consistently negligible for top-quark
tagging across all seeds, while vector-like channels are dominant but vary
substantially across independently trained seeds, consistent with the network
exploiting multiple representational pathways rather than a single fixed strategy.

These findings demonstrate that equivariance tests, linear probes
and grade ablations transfer directly to models with structured Lorentz
representations, suggesting they can guide validation and design of future
equivariant collider models.

\textbf{Relation to prior work.}
Recent interpretability studies of jet transformers have focused mainly on
attention maps for non-equivariant architectures.
\citet{wang2024interpreting} study \part{} attention, finding sparsity and
subjet structure, while \citet{legge2025attention} caution that attention sparsity does not by
itself explain model decisions. \citet{esmail2026iaformer} use attention
maps and CKA to analyze learned representations in IAFormer.
Linear probes were used in an early jet-tagging interpretability study by
\citet{cheng2019interpretability}, who found that individual N-subjettiness
values are more accessible than their ratios. We extend this probe-based
approach to transformer architectures and Lorentz-equivariant models and add
grade-aware interventions specific to geometric-algebra networks.
\lloca{}~\citep{spinner2025lloca} provides a comparison point between full
architectural equivariance and non-equivariant baselines.

The Appendix reports supporting diagnostics: attention maps, CKA layer
similarity~\citep{kornblith2019similarity}, full scalar- and multivector-probe
trajectories, grouped and subgroup-false ablation tables and details of the
combined-bootstrap uncertainty estimates used in the figures and tables.

\section{Background}
\label{sec:background}

\subsection{Geometric Algebra and the Lorentz Group}

The geometric algebra over Minkowski spacetime, with metric signature $(+,-,-,-)$,
is a $16$-dimensional associative algebra generated by four basis vectors
$e_0, e_1, e_2, e_3$ satisfying $e_0^2 = +1$, $e_i^2 = -1$ for $i=1,2,3$,
and $e_\mu e_\nu = -e_\nu e_\mu$ for $\mu \neq \nu$. Elements are called
\emph{multivectors} and decompose into five \emph{grades}:
\begin{itemize}[noitemsep,topsep=0pt,parsep=0pt,partopsep=0pt]
  \item \textbf{Grade 0} (scalar, 1 component): Lorentz-invariant by construction.
  \item \textbf{Grade 1} (vectors, 4 components): transform as 4-vectors under $\mathrm{SO}^+(1,3)$.
  \item \textbf{Grade 2} (bivectors, 6 components): oriented spacetime planes.
  \item \textbf{Grade 3} (trivectors, 4 components): Hodge-dual to G1 vectors via the pseudoscalar $I$.
  \item \textbf{Grade 4} (pseudoscalar, 1 component): parity-odd invariant.
\end{itemize}
We write grade $k$ as G$k$ throughout the paper; for example, G2 denotes
bivectors.

The geometric product of two multivectors mixes grades in a way determined
by the algebra; for instance, the product of two G2 elements produces
G0 and G4 components, while a G2 element times a G1 element produces G1
and G3 components.

\subsection{Lorentz-Equivariant Geometric Algebra Transformer}

\lgatr{}~\citep{spinner2024lgatr} adapts the Geometric Algebra Transformer
framework~\citep{brehmer2023gatr} from Euclidean to Minkowski spacetime.
It represents each jet constituent as a multivector token encoding its
4-momentum, supplemented by beam spurion tokens (a lightlike vector along
the beam axis and a time reference) that break the remaining boost symmetry
as a fixed reference frame. The network consists of $N_{\mathrm{blocks}}$
GATrBlocks, each applying equilinear transformations (linear maps that commute
with the Lorentz group action) and geometric attention over multivector tokens.
Each block produces multivector channels
$\hmv \in \mathbb{R}^{N \times C_{\mathrm{mv}} \times 16}$
and scalar channels $\hs \in \mathbb{R}^{N \times C_{\mathrm{s}}}$.

The output projection layer is equilinear: it maps grade $k$ inputs
to grade $k$ outputs only, with no grade mixing. Since the classification
logit is a scalar (G0), the output layer reads \emph{only} from the
G0 component of $\hmv$ at the final layer. This architectural constraint
implies that all non-scalar components at the final layer are architecturally
inaccessible to the output, regardless of their information content. We use this fact
explicitly in interpreting the layer-resolved ablations.

The scalar channels $\hs$ are Lorentz-invariant by construction.

By default \lgatr{} enforces equivariance to the connected, proper
orthochronous Lorentz subgroup $\mathrm{SO}^+(1,3)$. Under this subgroup,
G0 and G4 carry scalar-type representations, while G1 and G3 carry equivalent
vector-type representations via Hodge duality; the implementation therefore
uses the mixed groups G0+G4 and G1+G3. An alternative setting that keeps all
five grades strictly independent is studied as a side check in
Appendix~\ref{app:subgroup_false}.

\slim{}~\citep{petitjean2025slim} is a compact variant that retains only
scalars (G0) and 4-vectors (G1) internally, explicitly dropping the
outer product and G2, G3, and G4 channels. It matches \lgatr{}'s task AUC while
being $6\times$ faster~\citep{petitjean2025slim}.

\subsection{LLoCa-Transformer}

\lloca{}~\citep{spinner2025lloca} is designed to enforce Lorentz equivariance
through local canonicalization rather than through multivector-valued layers.
It maps each jet to a canonical Lorentz frame and applies a standard
transformer. Since the classification output is a Lorentz scalar, no inverse
frame map is required. By design \lloca{} is exactly Lorentz-equivariant: the canonicalization
is itself an equivariant operation, so the full pipeline inherits exact
equivariance. However, the canonicalization step is numerically more sensitive
than the architectural approach of \lgatr{}, which likely accounts for the
larger measured errors in our evaluation.
\lloca{} uses no multivector representation; its hidden state is a
flat vector, so grade decomposition does not apply directly.

\section{Experimental Setup}
\label{sec:setup}

\subsection{Dataset}

\textbf{TopTagging}~\citep{butter2018lorentz, kasieczka2019top} is a binary
jet tagging benchmark with 1.2M training, 400k validation and 400k test jets
at $\sqrt{s} = 14\,\mathrm{TeV}$. Signal jets arise from hadronic top quark
decays $t \to Wb \to qqb$ (three-pronged substructure); background jets are
QCD-initiated single-prong jets. Constituent particles are ordered by
transverse momentum $p_T$ (descending).

\subsection{Models}

We study five TopTagging models with comparable classification performance
(Table~\ref{tab:model_auc}).
\lgatr{}~\citep{spinner2024lgatr} has 12 GATrBlock layers with
$C_{\mathrm{mv}} = 16$ and $C_{\mathrm{s}} = 32$; \slim{}~\citep{petitjean2025slim}
has 12 LGATrSlimBlock layers with $C_v = 32$ and $C_s = 96$; and
\lloca{}~\citep{spinner2025lloca} has 12 transformer blocks with Lorentz
equivariance enforced by local canonicalization. These three models provide
the symmetry-aware comparison set.
We compare them to two non-equivariant baselines: \vanilla{}, a 10-block
standard transformer using the same general architecture family as \lgatr{},
and \part{}~\citep{qu2022particle}, a 10-layer kinematics-only
ParticleTransformer checkpoint. ParT augments self-attention with learned
pairwise physics biases, giving it a soft geometric prior without exact
equivariance.
All AUC values are 3-seed means.

\begin{table}[h]
  \vskip 0.05in
  \caption{Test AUC on TopTagging (mean $\pm$ std across 3 seeds).}
  \label{tab:model_auc}
  \vskip 0.05in
  \begin{center}
    \begin{small}
      \begin{tabular}{lc}
        \toprule
        Model & AUC \\
        \midrule
        \lgatr{}      & $0.9869 \pm 0.0001$ \\
        \slim{}       & $0.9867 \pm 0.0001$ \\
        \lloca{}      & $0.9867 \pm 0.0001$ \\
        \part{}       & $0.9857 \pm 0.0001$ \\
        \vanilla{}    & $0.9856 \pm 0.0001$ \\
        \bottomrule
      \end{tabular}
    \end{small}
  \end{center}
  \vskip -0.1in
\end{table}

\subsection{Probe Methods}

\textbf{Equivariance test.} For each model, we apply 5 random Lorentz transforms
to 200 test jets and measure the mean relative change in the output logit.
For \lgatr{} and \slim{}, the transform is applied to the full embedded input,
including beam-spurion/reference tokens, so the tested representation
transforms consistently.
We additionally perform a boost sweep with
$\gamma \in \{1.0, 1.25, 1.5, 1.75, 2.0, 2.5, 3.0, 4.0, 5.0\}$
to probe behavior from the identity transform through large boosts; the range
is capped at $\gamma \approx 5$,
motivated by the dataset's hard rapidity cut of $|\eta_j| < 2$, which
makes boosts well beyond $\gamma \sim 5$ poorly represented in the data. The output metric is
$|\ell(\Lambda x) - \ell(x)| / (|\ell(x)| + \varepsilon)$ with
$\varepsilon = 10^{-8}$. This relative-logit metric is a sensitive invariance
diagnostic, not a calibrated performance metric.

\textbf{Linear probes on $\hs$.} We extract scalar channel representations
at each layer for 10,000 TopTagging jets, then fit linear probes for 14
physics targets:
jet mass, jet multiplicity, N-subjettiness values
$\tau_1, \tau_2, \tau_3, \tau_{21}, \tau_{32}$~\citep{thaler2011njettiness},
top/QCD classification and particle-level $p_T$, $\eta$, $\phi$, $E$,
$p_T$-quartile and $\Delta R$. The main probe table and probe figures report
combined 95\% bootstrap intervals obtained by pooling draws across the 3
training seeds ($3 \times 200 = 600$ draws per target). Regression targets use Ridge regression;
binary and multi-class targets use logistic regression. For particle-level
probes, train/test separation is performed at the jet level.

\textbf{Grade decomposition.} We zero one grade group of $\hmv$ at all layers
simultaneously and measure the AUC drop (\emph{zero-grade ablation}),
and separately activate only one group while zeroing all others
(\emph{keep-only ablation}). Layer-resolved ablation zeros one grade group at
one layer at a time.
Under the default connected-subgroup setting (see Section~\ref{sec:background}),
L-GATr enforces equivariance to the proper orthochronous Lorentz subgroup, so scalar and
pseudoscalar components (G0+G4) may mix, as may vector and trivector
components (G1+G3); we therefore report three groups:
\emph{scalar-like} (G0+G4), \emph{vector-like} (G1+G3) and
\emph{bivector} (G2).
All ablations are applied to hidden multivector outputs after the corresponding
layer module, not to the raw input embedding.

\textbf{Geometric-algebra-invariant probes on $\hmv$.} We construct 832
Lorentz-invariant scalar features per token from $\hmv$ via the
geometric-algebra inner product $\langle \tilde{X} Y \rangle_0$ across
grade pairs, where $\tilde{\cdot}$ denotes reversion. Features include:
G0 and G4 direct components (32 features), G1 pairwise inner products
(136), G3 pairwise inner products (136), G1 $\times$ G3
cross terms (256) and G2 scalar and pseudoscalar invariants (136+136).
Implementation details are in Appendix~\ref{app:mv_probes_full}.

\section{Equivariance Validation}
\label{sec:equivariance}

\begin{figure}[t]
  \vskip 0.1in
  \begin{center}
    \includegraphics[width=\columnwidth]{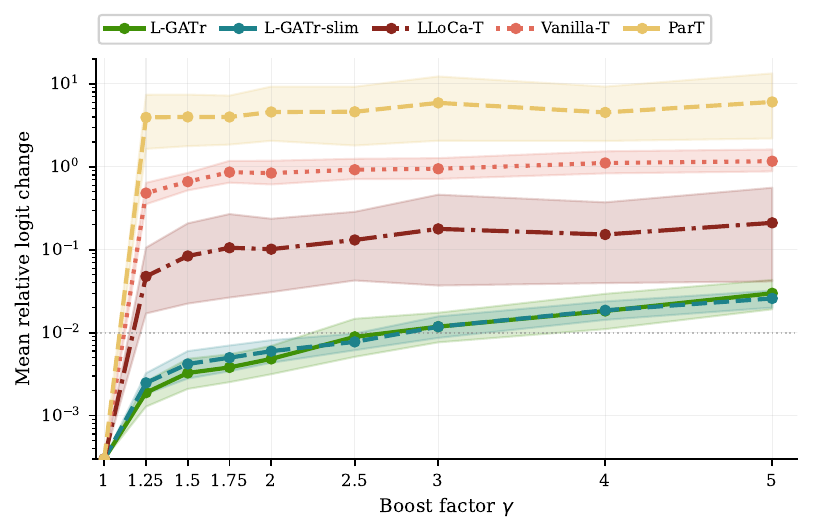}
    \caption{Mean relative logit change under pure boosts versus $\gamma$ across
      five models, using a sweep from $\gamma=1$ to $\gamma=5$. Shaded bands:
      combined 95\% bootstrap CIs pooled across 3 seeds. Lower is more equivariant.}
    \label{fig:equivariance}
  \end{center}
  \vskip -0.1in
\end{figure}

Figure~\ref{fig:equivariance} shows the mean relative logit change under pure
boosts across all five models, with combined 95\% bootstrap CIs pooled across
3 seeds; full combined-bootstrap intervals for random transforms are reported
in Appendix~\ref{app:equivariance_bootstrap}.
Under random Lorentz transforms, both \lgatr{} and \slim{} achieve logit errors
below 1\% (bootstrap medians $0.52\%$ and $0.39\%$ respectively), with
overlapping seed ranges, so the gap between the two equivariant models is
within training variability and should not be over-interpreted.
\lloca{} shows a larger measured error (${\approx}13\%$) than the
architectural models, consistent with its canonicalization being more
numerically sensitive than the architectural approach. \part{} and \vanilla{}
exhibit logit changes of ${\approx}334\%$ and ${\approx}120\%$ respectively,
roughly two to three orders of magnitude larger than the equivariant models.
We treat this experiment as validation rather than a main result: it confirms
that the trained models preserve the intended symmetries closely enough for the
representation-level analyses below.

Both \lgatr{} and \slim{} remain below $1\%$ logit error at moderate boosts
($\gamma \lesssim 2$). Their errors increase as the boost grows, consistent
with accumulated floating-point error at more extreme boosts. \lloca{} remains
much more stable than the non-equivariant baselines, but has wider and larger
measured errors than \lgatr{}/\slim{}, again reflecting greater numerical
sensitivity of the canonicalization. \part{} and \vanilla{} degrade sharply
even at moderate boosts.

\section{Physical Representations}
\label{sec:representations}

\subsection{Scalar Channel Probes}
\label{sec:hs_probes}

\begin{figure*}[t]
  \vskip 0.1in
  \begin{center}
    \includegraphics[width=0.97\textwidth]{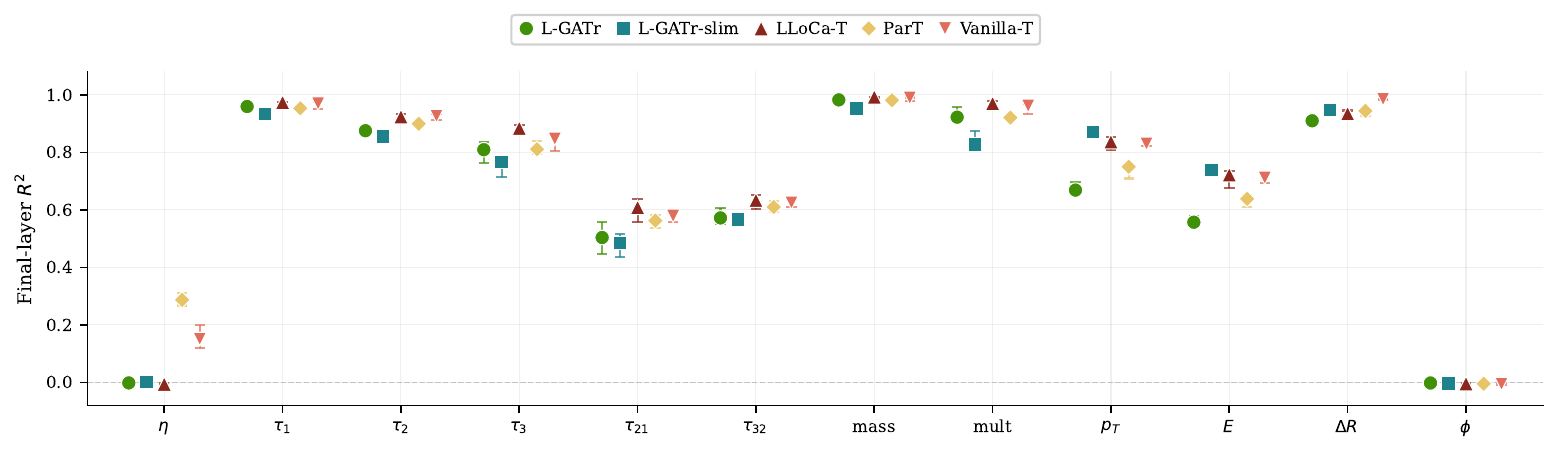}
    \caption{Final-layer linear probe scores for selected targets across five
      models. Points: combined-bootstrap medians; error bars: 95\% CIs pooled
      across 3 seeds. Targets shown: particle $\eta$, the N-subjettiness family,
      jet-level mass and multiplicity, then particle kinematics; particle $\phi$
      is placed last as it is near zero for all models. Top/QCD classification
      and particle $p_T$-rank quartile are reported in Table~\ref{tab:probes_full};
      full per-layer trajectories are in Appendix~\ref{app:uq}.}
    \label{fig:probe_bar}
  \end{center}
  \vskip -0.1in
\end{figure*}

Table~\ref{tab:probes_full} (Appendix~\ref{app:uq}) and
Figure~\ref{fig:probe_bar} report final-layer linear probe scores for all five
models as combined-bootstrap medians. Several consistent patterns emerge.

\textbf{Pseudorapidity $\eta$ as an invariance probe.}
All three symmetry-aware models suppress particle-level $\eta$ to approximately
zero at the final layer (\lgatr{}: $\bootval{-0.002}{-0.006,\,0.000}$;
\slim{}: $\bootval{0.000}{-0.005,\,0.005}$;
\lloca{}: $\bootval{-0.006}{-0.011,\,-0.002}$), while \vanilla{} retains
$\bootval{0.151}{0.119,\,0.198}$ and \part{} retains
$\bootval{0.287}{0.267,\,0.310}$. The particle-level $\phi$ probe is near zero
for all models, largely because inputs use $\Delta\phi$ relative to the jet axis
rather than absolute azimuth.
The $\eta$ result is a representation-level signature of the imposed symmetry:
in the architecturally equivariant models this absence is largely enforced by the scalar
channel construction, while \lloca{} shows a similar empirical pattern through
its canonicalization-based representation. The comparison to \vanilla{} and
\part{} is the non-trivial part: ordinary transformers retain linearly
accessible rapidity information.

\textbf{\lloca{} makes physics observables most linearly accessible.}
Despite its larger equivariance error, \lloca{} achieves the highest final-layer
probe scores for jet mass ($\bootval{0.993}{0.987,\,0.994}$), jet multiplicity
($\bootval{0.971}{0.957,\,0.977}$), $\tau_1$, $\tau_2$, $\tau_3$ and
$\tau_{21}$ ($\bootval{0.609}{0.557,\,0.637}$) of all five
models. This suggests that probe accessibility reflects
how linearly the model encodes physics: canonicalization may organize
representations in a way that is more linearly decodable in this probe basis,
despite its larger measured equivariance error.

\textbf{N-subjettiness and ratio probes.}
All models encode individual N-subjettiness values $\tau_1, \tau_2, \tau_3$
strongly ($R^2 \approx 0.84$--$0.97$), consistent with these being well-defined
jet substructure observables~\citep{thaler2011njettiness}. However, the ratio
$\tau_{21} = \tau_2 / \tau_1$ and the top-tagging-motivated ratio
$\tau_{32} = \tau_3 / \tau_2$ are substantially weaker for all models. This is
consistent with \citet{cheng2019interpretability}: networks encode the
building blocks $\tau_N$ more accessibly than their ratios, a pattern that
persists across all five architectures studied here.

\textbf{Full \lgatr{} makes jet-level observables more linearly accessible.}
Despite comparable task AUC, \lgatr{} achieves higher final-layer probe scores
than \slim{} for jet mass ($\bootval{0.983}{0.977,\,0.984}$ vs.\
$\bootval{0.953}{0.929,\,0.962}$), jet multiplicity
($\bootval{0.923}{0.907,\,0.957}$ vs.\
$\bootval{0.827}{0.816,\,0.874}$) and $\tau_3$
($\bootval{0.809}{0.764,\,0.837}$ vs.\
$\bootval{0.766}{0.714,\,0.779}$). This advantage is clearest for jet-level
substructure targets; several particle-level kinematic probes are higher for
\slim{}.

\textbf{\slim{} self-corrects input leakage.}
\slim{} takes $\Delta\eta$ as a scalar input feature and its particle-$\eta$
probe starts above zero at early layers before decaying to $\approx 0$ by the
final layer, consistent with progressively suppressing linearly accessible
non-invariant information in the probed scalar channels.

\begin{figure*}[t]
  \vskip 0.1in
  \begin{center}
    \includegraphics[width=0.97\textwidth]{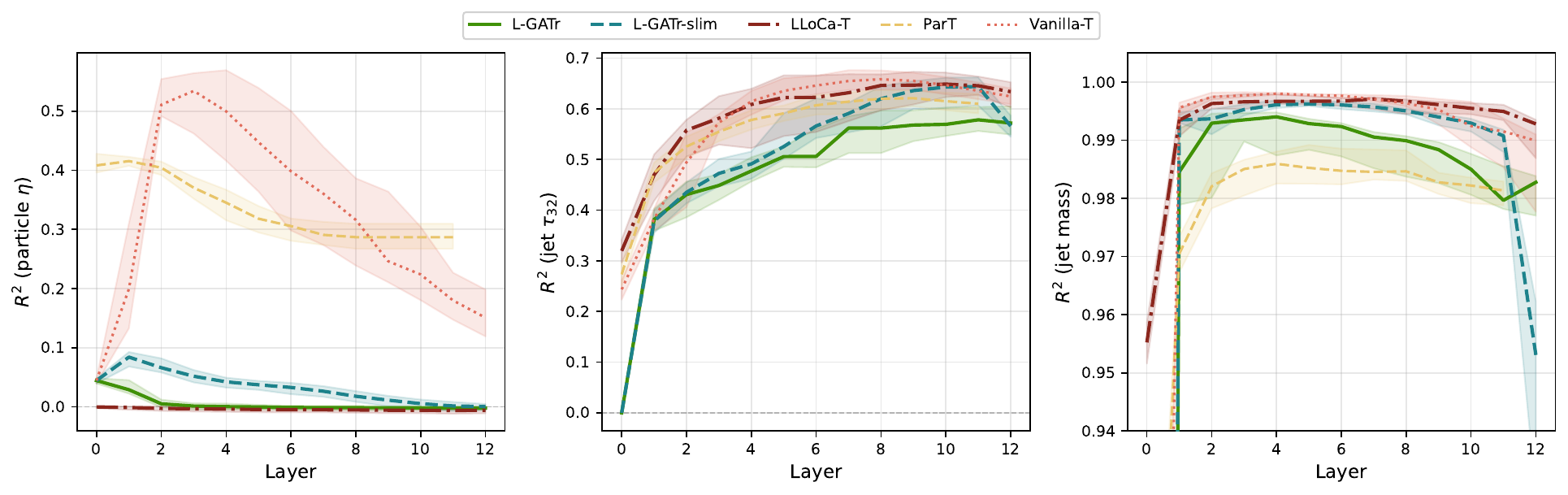}
    \caption{Scalar-channel probe trajectories across all five models. Shaded
      bands: combined 95\% bootstrap CIs pooled across 3 seeds.
      \textbf{Left}: particle $\eta$ — all three symmetry-aware models remain
      near zero throughout depth; \part{} and \vanilla{} retain substantial
      signal. \textbf{Centre}: $\tau_{32}$ is consistently harder to decode
      than individual $\tau_N$ values across all architectures.
      \textbf{Right}: jet mass — \lloca{} achieves the highest final-layer
      score. Full probe trajectories for all targets are in Appendix~\ref{app:uq}.}
    \label{fig:hs_summary}
  \end{center}
  \vskip -0.1in
\end{figure*}

Figure~\ref{fig:hs_summary} shows layerwise probe trajectories for $\eta$,
$\tau_{32}$, and jet mass. The $\eta$ suppression is visually immediate:
equivariant models remain near zero throughout depth, while \part{} and
\vanilla{} retain substantial non-zero signal.

\subsection{Multivector Invariant Probes}
\label{sec:mv_probes}

We fit linear probes on the 832 Lorentz-invariant features from $\hmv$ at each
layer of \lgatr{} (Section~\ref{sec:setup}). Two findings inform the grade
discussion below.

\textbf{Multiple usable pathways in the final layer.}
At the final layer, probes on scalar-like invariants (G0+G4 components,
32 features) and vector-like invariants (G1+G3 inner products and cross terms,
528 features) both approach full-model classification AUC:
scalar-like achieves $\approx 0.986$ and vector-like achieves
$\approx 0.982$--$0.986$ under the combined-bootstrap summary, matching the task AUC.
This is consistent with \lgatr{} encoding discriminative information through
both the vector-like (G1+G3) pathway and the scalar-like (G0+G4) pathway,
as reflected also in the high cross-seed variance of
vector-like ablations in Section~\ref{sec:grade}.

\textbf{$\eta$ is absent from invariant features.}
The particle $\eta$ probe on all 832 invariant features gives $R^2 \approx -0.025$
(seed mean, negative for all seeds), confirming that the Lorentz-invariant
features carry no frame-dependent rapidity information. The particle $\phi$
probe is near zero across all layers ($R^2 \approx 0$).

\textbf{$\tau_{21}$ more accessible from $\hmv$ than $\hs$.}
The full 832-feature probe achieves $R^2 \approx 0.648$ for $\tau_{21}$
at the final layer, compared to $0.504$ from $\hs$ alone (Table~\ref{tab:probes_full}).
The ratio appears to be encoded in the inter-grade structure of $\hmv$ but
less linearly accessible from the scalar channels. Full per-layer trajectories
are in Appendix~\ref{app:mv_probes_full}.

\section{Grade Structure}
\label{sec:grade}

\subsection{Grade Ablations: Grouped Pathways and Bivector Redundancy}
\label{sec:grade_ablations}

Table~\ref{tab:grade_decomp} reports grouped grade-ablation results for
\lgatr{} on TopTagging, summarized across three seeds. Figure~\ref{fig:grade_layer}
plots the same zero-grade results and shows how the zero-grade intervention
changes with depth.

\textbf{Bivectors are not load-bearing in this setting.}
Zeroing all bivector (G2) channels at every layer reduces AUC by only
$\bootval{0.001}{0.000,\,0.005}$, the smallest effect of any grade
group and consistent with zero. This finding is seed-robust: no seed shows
G2 ablation impact above 0.005. This does not imply that bivectors are never
useful in L-GATr; rather, for these TopTagging checkpoints the task information
is recoverable through the scalar-like and vector-like pathways under the
default connected-subgroup parameterization.

\textbf{Vector-like channels are dominant but seed-variable.}
Zeroing vector-like (G1+G3) channels gives
$\dauc=\bootval{0.239}{0.212,\,0.816}$, the largest zero-ablation effect,
but with very high cross-seed variance
(seed values: $0.809$, $0.239$, $0.219$). This variability suggests that
different training runs may allocate task information differently between the
vector-like (G1+G3) pathway and the scalar-like (G0+G4) pathway, as discussed in
Section~\ref{sec:mv_probes}. More seeds would be needed to characterize this
pathway preference definitively.

\textbf{Scalar-like channels remain competitive.}
Zeroing scalar-like (G0+G4) channels gives
$\dauc=\bootval{0.118}{0.035,\,0.173}$. The keep-only view is complementary:
keeping only scalar-like channels gives $\dauc=\bootval{0.239}{0.212,\,0.816}$,
matching the vector-like zero ablation. Taken together with the invariant-probe
results, this supports redundancy between scalar-like and vector-like pathways.

\textbf{Note on keep-only bivector.}
Keeping only G2 channels gives $\dauc=\bootval{0.337}{0.095,\,0.397}$
(Table~\ref{tab:grade_decomp}),
which may appear in tension with the G2$\approx$0 zero-ablation result.
The keep-only intervention does not isolate standalone bivector capacity:
the scalar branch remains available to the readout, so the remaining
performance can still use scalar information. We therefore interpret the
near-zero G2 zero-ablation as the cleaner evidence that bivectors are not
load-bearing here.

\begin{table}[t]
  \caption{Grade decomposition ablations for \lgatr{} on TopTagging
    (default connected-subgroup setting). Entries give combined-bootstrap
    medians with asymmetric 95\% intervals; the left panel of
    Figure~\ref{fig:grade_layer} plots the zero-grade column.}
  \label{tab:grade_decomp}
  \vskip 0.1in
  \begin{center}
    \resizebox{\columnwidth}{!}{%
      \begin{sc}
        \renewcommand{\arraystretch}{1.40}
        \begin{tabular}{lcc}
          \toprule
          Group & Zero $\dauc$ & Keep-only $\dauc$ \\
          \midrule
          scalar-like (G0+G4)  & $\asymci{0.118}{0.056}{0.082}$  & $\asymci{0.239}{0.577}{0.027}$ \\
          vector-like (G1+G3)  & $\asymci{0.239}{0.577}{0.027}$  & $\asymci{0.149}{0.014}{0.099}$ \\
          bivector (G2) & $\asymci{0.001}{0.004}{0.001}$ & $\asymci{0.337}{0.060}{0.242}$ \\
          \bottomrule
        \end{tabular}
      \end{sc}%
    }
  \end{center}
  \vskip -0.1in
\end{table}

\subsection{Layer-Resolved Ablation}
\label{sec:grade_layer}

\begin{figure*}[t]
  \vskip 0.1in
  \begin{center}
    \includegraphics[width=\textwidth]{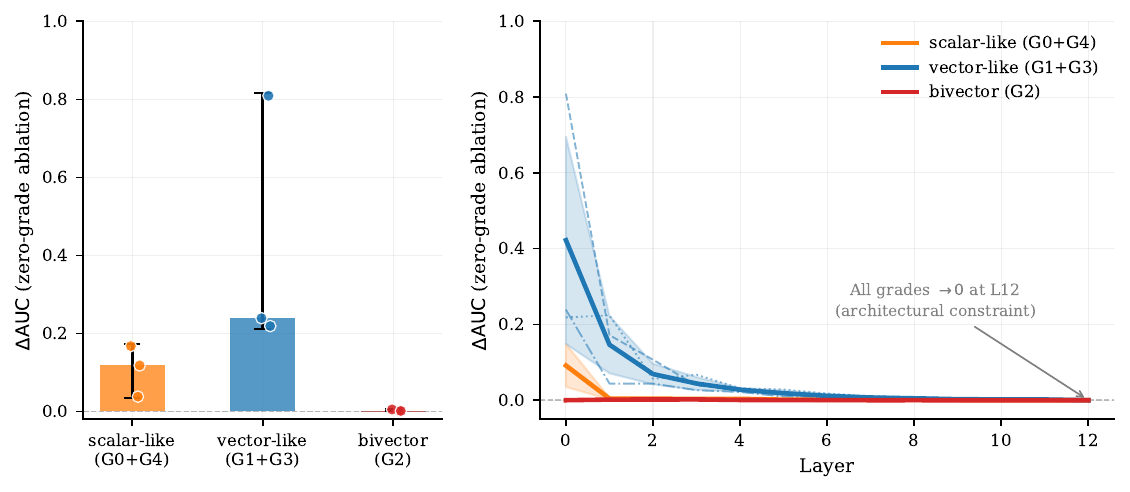}
    \caption{Grade decomposition for \lgatr{} on TopTagging.
      \textbf{Left}: global zero-grade ablation $\dauc$ for each grade group.
      Bars show combined 95\% bootstrap CIs pooled across 3 seeds; dots show per-seed values.
      Bivector (G2) is consistently negligible; vector-like (G1+G3) is dominant
      but highly seed-variable, reflecting multiple usable pathways.
      \textbf{Right}: layer-resolved zero-grade ablation (seed mean $\pm$ std;
      thin lines show per-seed vector-like curves). Vector-like impact is largest
      at the input stage and decays monotonically; bivector is negligible at every layer.
      All grade groups converge to $\dauc \approx 0$ at the final layer by architectural necessity
      (grade-preserving output projection).}
    \label{fig:grade_layer}
  \end{center}
  \vskip -0.1in
\end{figure*}

Figure~\ref{fig:grade_layer} shows the global and layer-resolved grade ablations
for \lgatr{} on TopTagging. Several patterns are clear.

Vector-like ablation impact is largest at the input stage
(mean $\dauc = 0.422$) and decays monotonically with depth. Scalar-like
impact is smaller throughout ($\dauc = 0.091$ at the input stage, near zero
after the first block).
Bivector ablation is negligible at every individual layer ($\dauc < 0.002$
at all depths). All grade-group ablations converge to $\dauc \approx 0$ at the final layer by
architectural necessity: the grade-preserving output projection can only
read G0 signal from the final layer, regardless of what other grades
encode. Seed 1001 shows substantially larger input-stage vector-like impact ($0.809$)
than seeds 1002/1003 (${\approx}0.28$), consistent with the seed-variable
global ablation result and the multiple-pathway hypothesis.

\subsection{Training without Bivectors}
\label{sec:bivector_false}

To verify that the G2 $\approx$ 0 finding is not an artifact of the trained
network suppressing bivectors post-hoc, we also study a model trained with
bivectors architecturally disabled (G2 channel zeroed during training).
This model achieves baseline AUC $0.9865 \pm 0.0002$, statistically
indistinguishable from the standard \lgatr{} ($0.9867 \pm 0.0003$). Its
vector-like $\dauc$ ($\bootval{0.216}{0.078,\,0.407}$) is comparable to seeds 1002/1003
of the standard model. Thus, architecturally removing G2 during training has
negligible effect on task performance. Full grade ablations for this model are in
Appendix~\ref{app:biv_false}. A complementary check with the
alternative subgroup setting (five independent grades rather than three mixed
groups) likewise finds G2 negligible and further shows that parity-odd grades
(G3, G4) are not load-bearing; full tables are in
Appendix~\ref{app:subgroup_false}.

\subsection{\slim{} Contrast}
\label{sec:slim_contrast}

\slim{} retains only G0 scalars and G1 4-vectors and has no G2, G3, or G4
channels. Its ablation pattern differs sharply from \lgatr{}: zeroing all scalar
channels reduces AUC to $0.500$ ($\dauc = \bootval{0.486}{0.485,\,0.488}$), a complete loss
of discriminative power, while zeroing the 4-vector channels gives
$\dauc = \bootval{0.585}{0.559,\,0.695}$. Full component-level ablations are in
Appendix~\ref{app:slim_grade}. The scalar branch is indispensable for \slim{}
because the readout is the global scalar token directly.

In contrast, \lgatr{} shows a more distributed pattern: vector-like channels
are the dominant but seed-variable ablation target, scalar-like channels remain
competitive, and the higher-order grades omitted by \slim{} do not appear
necessary for TopTagging.
Both architectures achieve comparable task AUC, suggesting that grade structure
reflects representational strategy rather than task requirement.

\section{Discussion and Conclusion}
\label{sec:discussion}

We have shown that equivariance tests, scalar-channel probes, grade
decomposition and geometric-algebra-invariant probes expose physically
structured internal representations in Lorentz-equivariant jet taggers. The
output-level checks validate that the symmetry constraints are active in the
trained models, while the probe results show how this symmetry appears inside
the representations: symmetry-aware models suppress linearly accessible
pseudorapidity while retaining strong access to jet mass and N-subjettiness.

\textbf{Probe accessibility and task use are distinct.}
\lloca{} achieves the highest linear probe scores for most physics targets,
despite larger measured numerical errors in the equivariance test. This suggests
that linear decodability is best interpreted as a property of representation
organization, not as a direct measure of task utility. The contrast between
\lgatr{} and \lloca{} is therefore informative: both suppress $\eta$, but their
internal representations expose different physics observables to simple linear
readouts.

\textbf{Grade structure reflects representational strategy.}
For \lgatr{}, the clearest grade result is negative but useful: G2 bivectors are
not load-bearing for these TopTagging checkpoints. They are negligible under
zero-ablation, remain negligible in the five-grade side study and can be removed
during training without degrading AUC. The more open result is the seed-variable
importance of vector-like channels. Together with the invariant-probe evidence,
this suggests that independently trained \lgatr{} models may distribute task
information differently across scalar-like and vector-like pathways, although
more seeds are needed to characterize that preference.

\textbf{Limitations.}
Linear probes measure accessibility of information, not necessarily how the
model uses it internally. Grade ablations are interventions on trained networks
and may introduce distribution shift, especially for keep-only settings. The
bivector result is consistent across 3 independent runs and the training-time
G2-disabled check, but the seed-variable vector-like pathway still warrants
additional trainings. Our analysis is limited to TopTagging.

\textbf{Future directions.}
A fuller JetClass analysis, including per-class grade structure and $\hs$
probes, would test whether grade patterns are task-dependent. Adapting
mechanistic interpretability methods such as activation patching, circuit
analysis or sparse autoencoders to geometric-algebra representations is an open
opportunity. Probing whether \lloca{}'s high linear-probe accessibility
translates to more interpretable intermediate representations is another
promising direction.

\section*{Acknowledgements}

We thank the anonymous reviewers for their helpful comments. 
We gratefully acknowledge the computational resources provided by BITS Pilani that made this work possible.

\section*{Impact Statement}

This paper presents work whose goal is to advance the interpretability of
symmetry-constrained neural networks for particle physics applications.
We do not anticipate specific negative societal consequences from this work.

\bibliography{references}

\begin{thebibliography}{13}
\providecommand{\natexlab}[1]{#1}
\providecommand{\url}[1]{\texttt{#1}}
\expandafter\ifx\csname urlstyle\endcsname\relax
  \providecommand{\doi}[1]{doi: #1}\else
  \providecommand{\doi}{doi: \begingroup \urlstyle{rm}\Url}\fi

\bibitem[Brehmer et~al.(2023)Brehmer, de~Haan, Behrends, and
  Cohen]{brehmer2023gatr}
Brehmer, J., de~Haan, P., Behrends, S., and Cohen, T.
\newblock {Geometric Algebra Transformer}.
\newblock In \emph{Advances in Neural Information Processing Systems},
  volume~36, 2023.
\newblock arXiv:2305.18415.

\bibitem[Butter et~al.(2018)Butter, Kasieczka, Plehn, and
  Russell]{butter2018lorentz}
Butter, A., Kasieczka, G., Plehn, T., and Russell, M.
\newblock {Deep-learned Top Tagging with a Lorentz Layer}.
\newblock \emph{SciPost Physics}, 5:\penalty0 028, 2018.
\newblock arXiv:1707.08966.

\bibitem[Cheng(2019)]{cheng2019interpretability}
Cheng, T.
\newblock {Interpretability Study on Deep Learning for Jet Physics at the Large
  Hadron Collider}.
\newblock In \emph{Machine Learning and the Physical Sciences Workshop,
  NeurIPS}, 2019.
\newblock arXiv:1911.01872.

\bibitem[Esmail et~al.(2026)Esmail, Hammad, and Nojiri]{esmail2026iaformer}
Esmail, W., Hammad, A., and Nojiri, M.
\newblock {IAFormer: Interaction-Aware Transformer network for collider data
  analysis}.
\newblock \emph{SciPost Physics}, 20:\penalty0 108, 2026.
\newblock arXiv:2505.03258.

\bibitem[Kasieczka et~al.(2019)Kasieczka, Plehn, Butter, Cranmer, Debnath,
  Dillon, et~al.]{kasieczka2019top}
Kasieczka, G., Plehn, T., Butter, A., Cranmer, K., Debnath, D., Dillon, B.~M.,
  et~al.
\newblock {The Machine Learning Landscape of Top Taggers}.
\newblock \emph{SciPost Physics}, 7:\penalty0 014, 2019.
\newblock arXiv:1902.09914.

\bibitem[Kornblith et~al.(2019)Kornblith, Norouzi, Lee, and
  Hinton]{kornblith2019similarity}
Kornblith, S., Norouzi, M., Lee, H., and Hinton, G.
\newblock {Similarity of Neural Network Representations Revisited}.
\newblock In \emph{Proceedings of the 36th International Conference on Machine
  Learning}, volume~97 of \emph{Proceedings of Machine Learning Research}, pp.\
   3519--3529, 2019.
\newblock arXiv:1905.00414.

\bibitem[Legge et~al.(2025)Legge, Wang, Ortiz, Limouzi, Zhao, Gandrakota,
  Khoda, Ngadiuba, Duarte, and Cavanaugh]{legge2025attention}
Legge, T., Wang, A., Ortiz, J., Limouzi, V., Zhao, Z., Gandrakota, A., Khoda,
  E.~E., Ngadiuba, J., Duarte, J., and Cavanaugh, R.
\newblock {Why Is Attention Sparse In Particle Transformer?}
\newblock In \emph{Machine Learning and the Physical Sciences Workshop,
  NeurIPS}, 2025.
\newblock arXiv:2512.00210.

\bibitem[Petitjean et~al.(2025)Petitjean, Plehn, Spinner, and
  K{\"o}the]{petitjean2025slim}
Petitjean, A., Plehn, T., Spinner, J., and K{\"o}the, U.
\newblock {Economical Jet Taggers -- Equivariant, Slim, and Quantized}.
\newblock \emph{arXiv preprint arXiv:2512.17011}, 2025.
\newblock arXiv:2512.17011.

\bibitem[Qu et~al.(2022)Qu, Li, and Qian]{qu2022particle}
Qu, H., Li, C., and Qian, S.
\newblock {Particle Transformer for Jet Tagging}.
\newblock In \emph{Proceedings of the 39th International Conference on Machine
  Learning}, volume 162 of \emph{Proceedings of Machine Learning Research},
  pp.\  18281--18292, 2022.
\newblock arXiv:2202.03772.

\bibitem[Spinner et~al.(2024)Spinner, Bres{\'o}, de~Haan, Plehn, Thaler, and
  Brehmer]{spinner2024lgatr}
Spinner, J., Bres{\'o}, V., de~Haan, P., Plehn, T., Thaler, J., and Brehmer, J.
\newblock {Lorentz-Equivariant Geometric Algebra Transformers for High-Energy
  Physics}.
\newblock In \emph{Advances in Neural Information Processing Systems},
  volume~38, 2024.
\newblock arXiv:2405.14806.

\bibitem[Spinner et~al.(2025)Spinner, Favaro, Lippmann, Pitz, Gerhartz, Plehn,
  and Hamprecht]{spinner2025lloca}
Spinner, J., Favaro, L., Lippmann, P., Pitz, S., Gerhartz, G., Plehn, T., and
  Hamprecht, F.~A.
\newblock {Lorentz Local Canonicalization: How to Make Any Network
  Lorentz-Equivariant}.
\newblock In \emph{Advances in Neural Information Processing Systems}, 2025.
\newblock arXiv:2505.20280.

\bibitem[Thaler \& Van~Tilburg(2011)Thaler and
  Van~Tilburg]{thaler2011njettiness}
Thaler, J. and Van~Tilburg, K.
\newblock {Identifying Boosted Objects with N-subjettiness}.
\newblock \emph{Journal of High Energy Physics}, 2011:\penalty0 015, 2011.
\newblock arXiv:1011.2268.

\bibitem[Wang et~al.(2024)Wang, Gandrakota, Ngadiuba, Sahu, Bhatnagar, Khoda,
  and Duarte]{wang2024interpreting}
Wang, A., Gandrakota, A., Ngadiuba, J., Sahu, V., Bhatnagar, P., Khoda, E.~E.,
  and Duarte, J.
\newblock {Interpreting Transformers for Jet Tagging}.
\newblock In \emph{Machine Learning and the Physical Sciences Workshop,
  NeurIPS}, 2024.
\newblock arXiv:2412.03673.

\end{thebibliography}
\bibliographystyle{icml2026}

\clearpage
\appendix
\onecolumn

\section*{Appendix}

\FloatBarrier
\section{Attention Maps}
\label{app:attention}

\begin{figure}[h!]
  \begin{center}
    \includegraphics[width=0.85\textwidth]{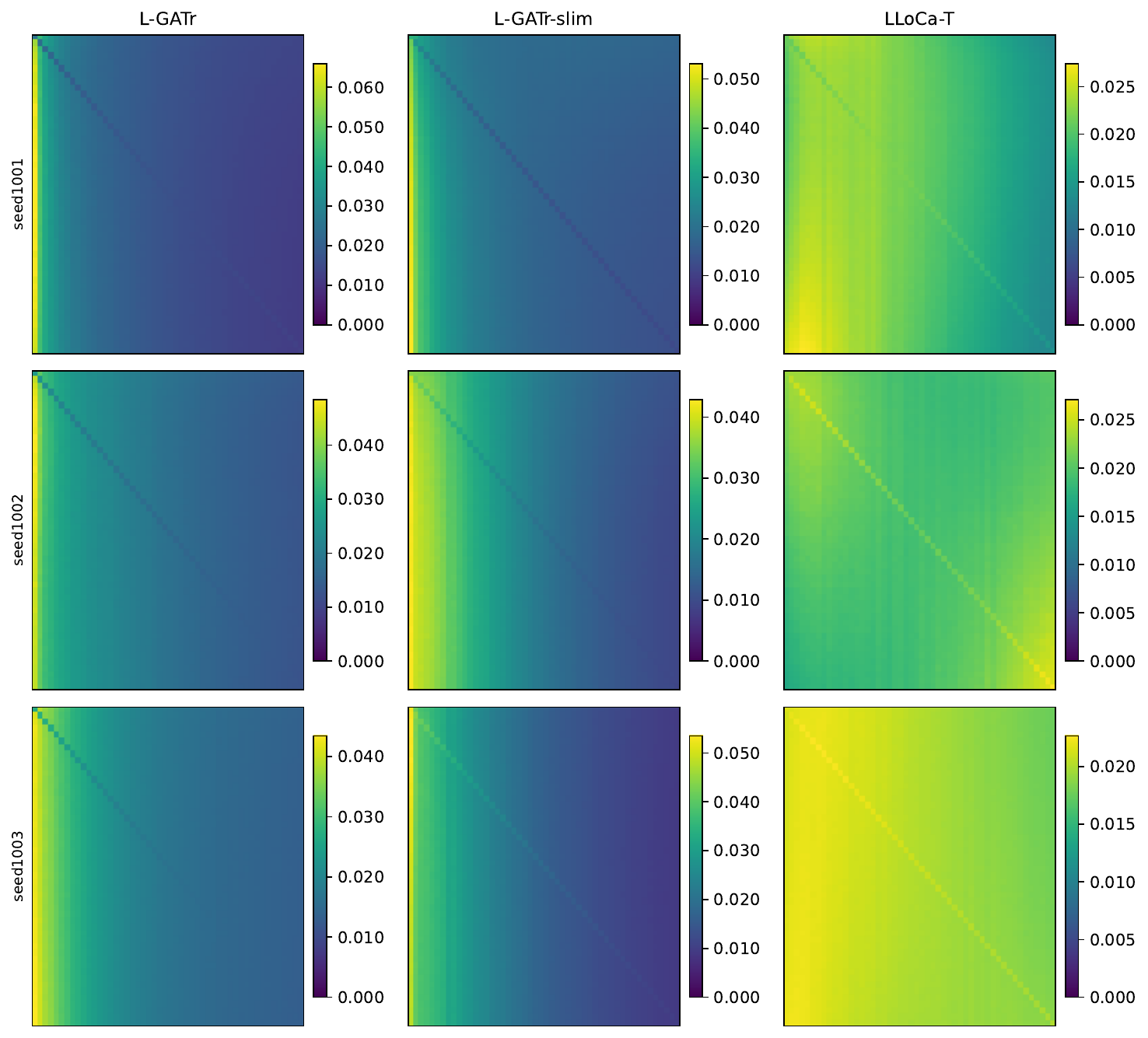}
    \caption{Overview: head-averaged mean final-layer attention maps for
      \lgatr{} (left), \slim{} (centre) and \lloca{} (right), across all
      three training seeds (rows). \lgatr{} and \slim{} both show strong attention to the
      leading-$p_T$ constituent; \lloca{} displays a more diffuse pattern.}
    \label{fig:attn_overview}
  \end{center}
\end{figure}

\begin{figure}[h!]
  \begin{center}
    \includegraphics[width=\textwidth]{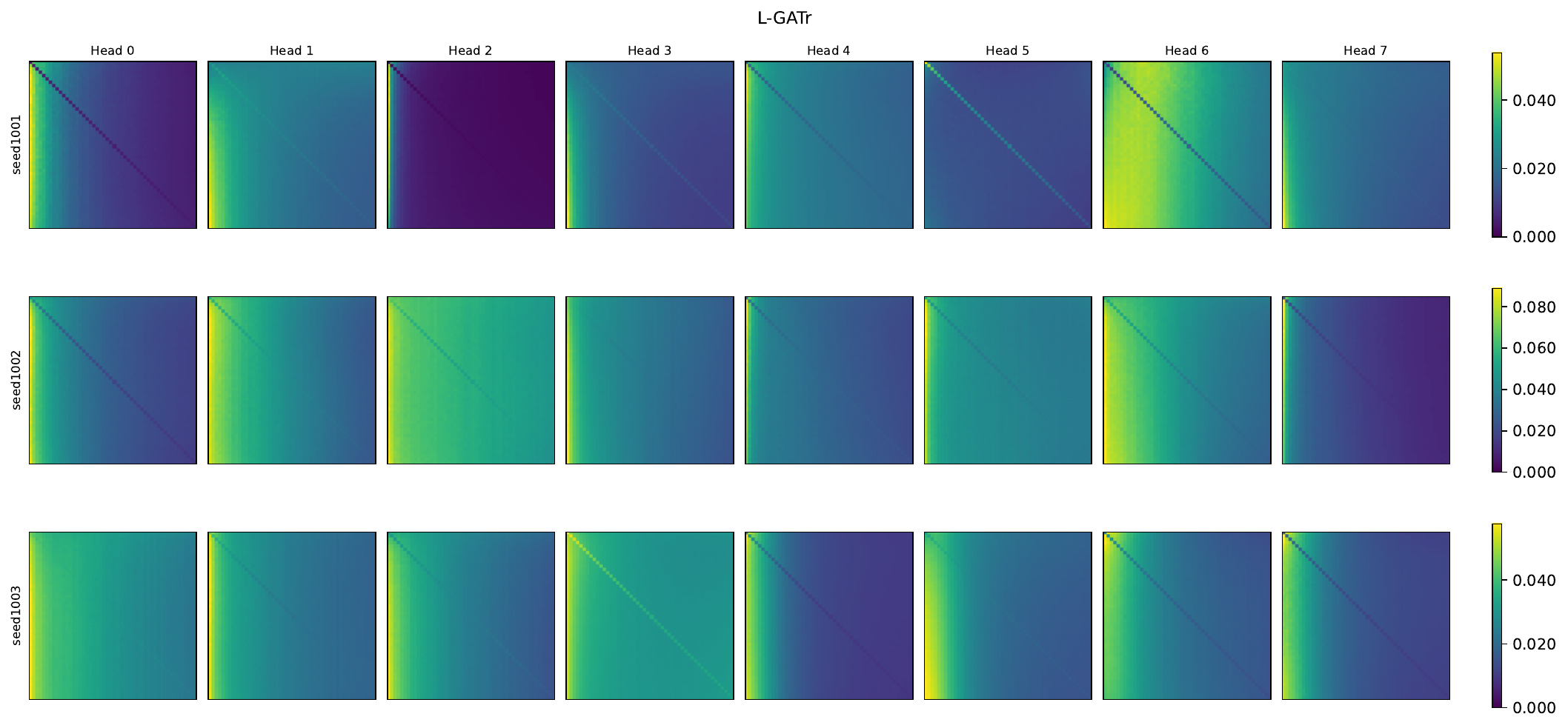}
    \caption{Individual attention heads (0--7) for \lgatr{}, across all
      three training seeds (rows). Head~2 shows an especially sharp
      particle-0 focus; other heads display varying degrees of leading-$p_T$
      attention.}
    \label{fig:attn_lgatr}
  \end{center}
\end{figure}

\begin{figure}[h!]
  \begin{center}
    \includegraphics[width=\textwidth]{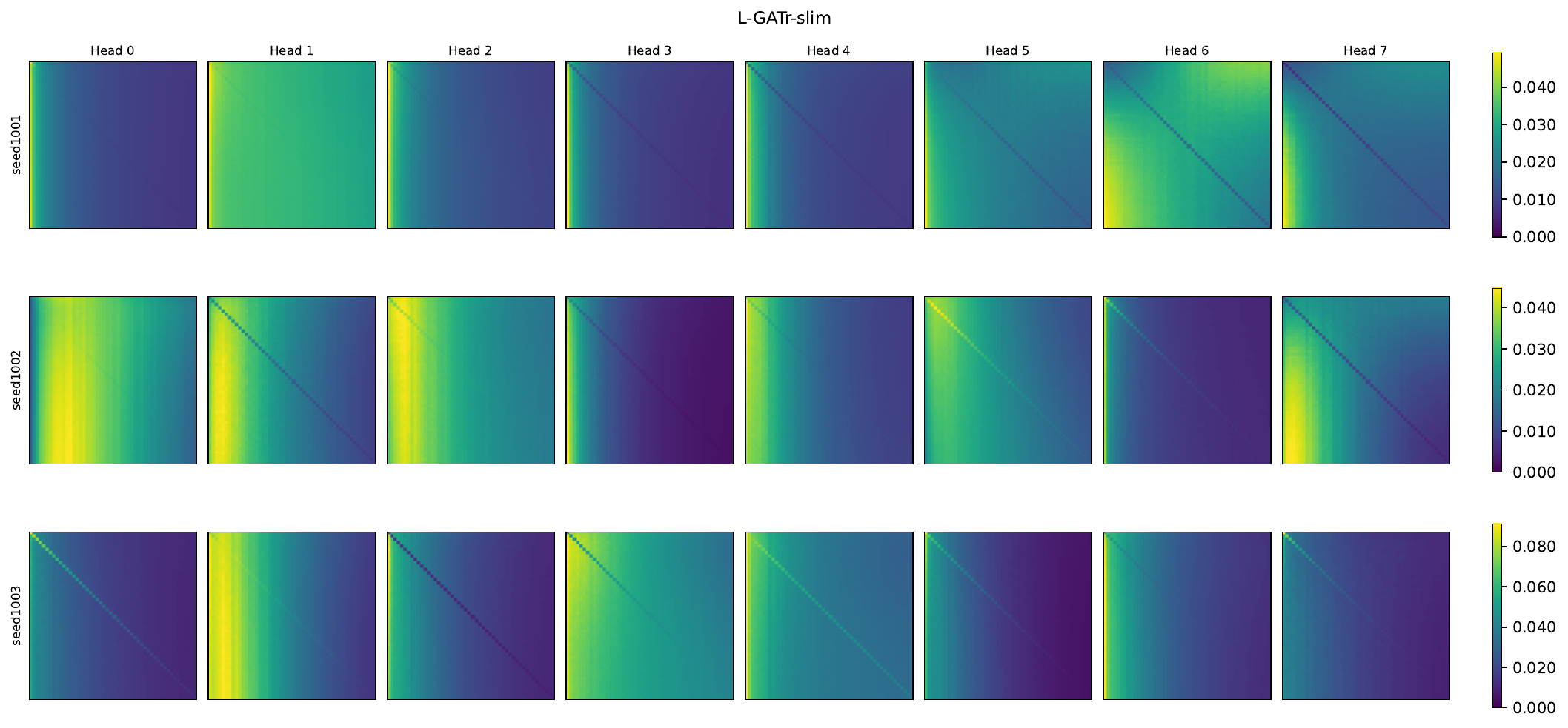}
    \caption{Individual attention heads (0--7) for \slim{}, across all
      three training seeds (rows). Pattern is broadly similar to \lgatr{}
      but with a somewhat broader spread across constituents.}
    \label{fig:attn_slim}
  \end{center}
\end{figure}

\begin{figure}[h!]
  \begin{center}
    \includegraphics[width=\textwidth]{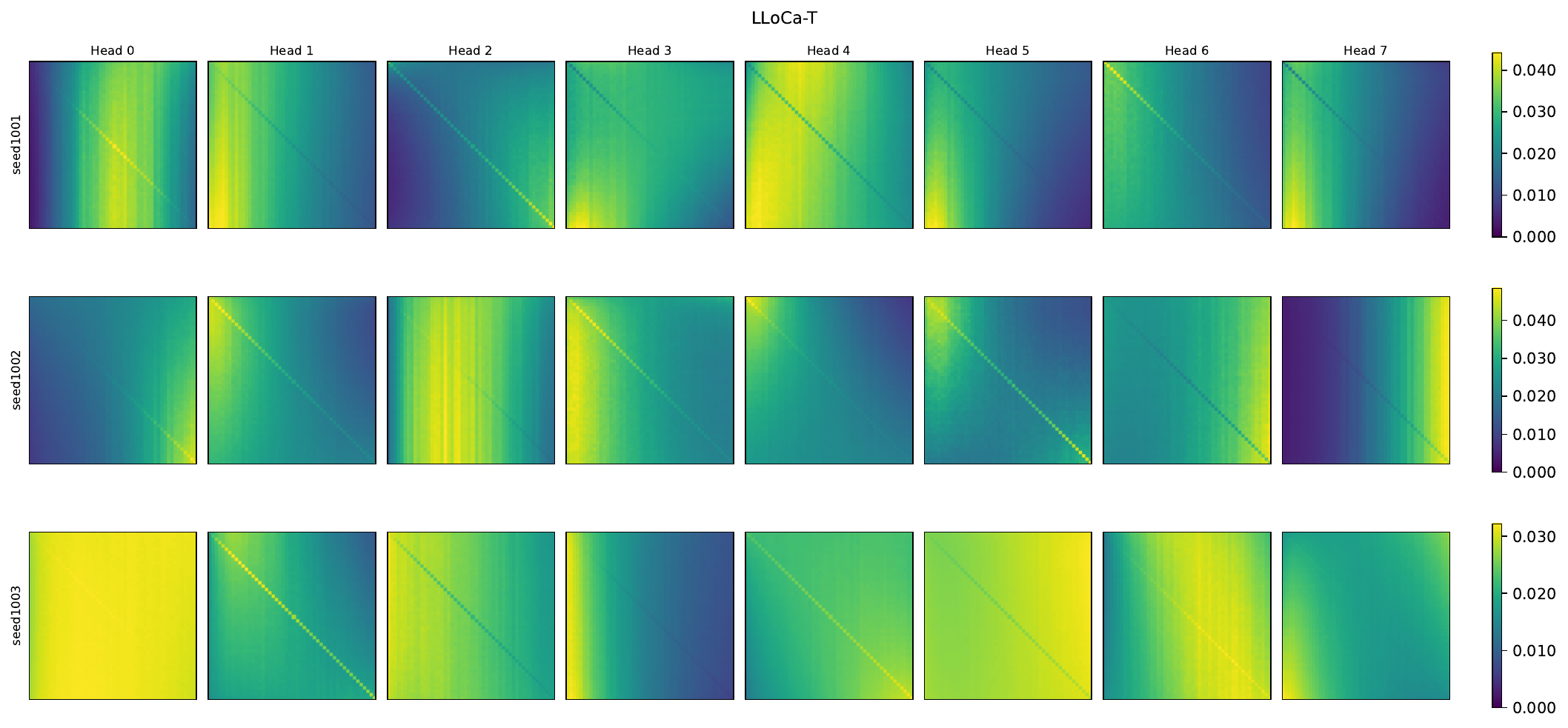}
    \caption{Individual attention heads (0--7) for \lloca{}, across all
      three training seeds (rows). Attention patterns are more spatially
      diffuse than \lgatr{}/\slim{}, reflecting the different internal
      structure of canonicalization-based equivariance.}
    \label{fig:attn_lloca}
  \end{center}
\end{figure}

Figures~\ref{fig:attn_overview}--\ref{fig:attn_lloca} show mean final-layer
attention maps on TopTagging for all three Lorentz-aware models (\lgatr{},
\slim{}, \lloca{}), truncated to the top-50 constituents by $p_T$.
Figure~\ref{fig:attn_overview} gives an overview of the head-averaged maps
across all three models and all three training seeds.
Figures~\ref{fig:attn_lgatr}, \ref{fig:attn_slim} and \ref{fig:attn_lloca}
break out all 8 individual attention heads per model.

Both \lgatr{} and \slim{} show a strong left-column pattern at position
$j=0$ in most heads, indicating that many heads attend heavily to the
highest-$p_T$ constituent. \lgatr{} tends to have sharper particle-0 focus in
several heads (Head~2 across all seeds), while \slim{} shows a somewhat
broader spread. \lloca{} displays a qualitatively different pattern: the
head-averaged map is more diffuse and lacks the sharp left-column feature,
reflecting the different representational strategy of canonicalization-based
equivariance. These patterns are consistent across all three seeds.

\FloatBarrier
\section{CKA Representation Similarity}
\label{app:additional}

We compute linear Centered Kernel Alignment
(CKA~\citep{kornblith2019similarity}) on scalar-channel representations as a
supporting diagnostic for how representations evolve with depth. Unlike the
probe and ablation analyses in the main text, these plots are descriptive:
they show where layers become similar or reorganise, but they do not by
themselves identify which physics observables are encoded.

\begin{figure}[h!]
  \begin{center}
    \includegraphics[width=\textwidth]{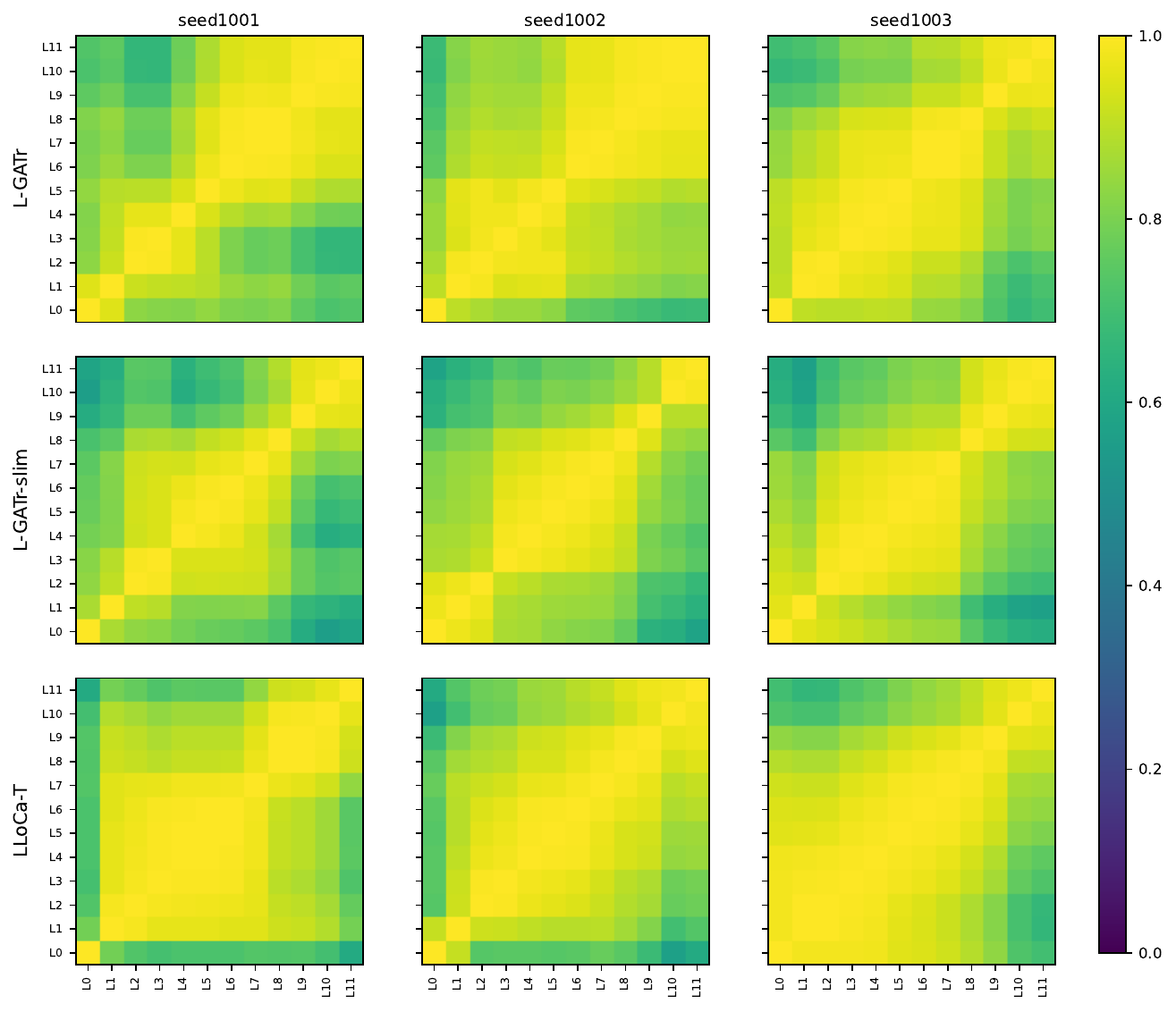}
    \caption{Linear CKA between scalar-channel representations at all
      layer pairs (L0--L11) for \lgatr{} (top row), \slim{} (middle row)
      and \lloca{} (bottom row), across all three training seeds (columns).
      Axes: L0 at bottom-left, L11 at top-right. Colorscale shared across
      all panels $[0, 1]$.}
    \label{fig:cka}
  \end{center}
\end{figure}

We compute linear CKA between scalar channel representations
$\hs$ at all pairs of layers, for \lgatr{}, \slim{} and \lloca{}
(5,000 jets, optimised Gram matrix approach). The matrices use transformer
block outputs L0--L11 (12 layers), with axes oriented so that
$(L0, L0)$ is at the bottom-left and $(L11, L11)$ at the top-right.
Figure~\ref{fig:cka} shows the $12 \times 12$ CKA matrices for all three
models and all three training seeds.
All three models show high adjacent-layer similarity and lower, but still
substantial, early--late similarity. Thus the scalar representations evolve
gradually with depth rather than undergoing a sharp reorganization at a single
layer; \lloca{} and \lgatr{} are slightly more globally self-similar than
\slim{} in these CKA diagnostics.

\FloatBarrier
\section{Scalar Channel Probes}
\label{app:uq}

\textbf{Probe training details.}
For the TopTagging $\hs$ probes, we work with 10,000 jets and fit probes at
every layer. Regression targets use Ridge regression with
$\alpha = 1.0$; binary and multi-class targets use logistic regression with
L2 regularization (\texttt{lbfgs}, $C = 1.0$). For particle-level targets,
train/test split is defined at the jet level.

\textbf{Scalar-probe uncertainty.}
For scalar-channel probes, shaded bands and tabulated intervals use combined
95\% bootstrap intervals obtained by pooling bootstrap draws across the 3
independently trained seeds. Each seed contributes 200 bootstrap draws per
model, target and layer, for 600 pooled draws per reported interval. This
convention summarizes both finite-sample variation within a checkpoint and
training-seed variation across independently trained checkpoints.

\begin{table}[H]
  \caption{Full linear probe uncertainty summary at the final layer for all five models.
    Format: combined-bootstrap median [95\% CI]. \textbf{Part $\eta$} row highlighted:
    $\approx 0$ for all symmetry-aware models (L-GATr, slim, LLoCa-T),
    non-zero for non-equivariant models.}
  \label{tab:probes_full}
  \vskip 0.1in
  \begin{center}
    \resizebox{\textwidth}{!}{%
      \begin{tabular}{lcccccc}
        \toprule
        Probe & \lgatr{} & \slim{} & \lloca{} & \vanilla{} & \part{} & Metric \\
        \midrule
        top/QCD        & $\bootci{0.986}{0.984,\,0.989}$ & $\bootci{0.986}{0.984,\,0.989}$ & $\bootci{0.986}{0.983,\,0.988}$ & $\bootci{0.985}{0.982,\,0.987}$ & $\bootci{0.984}{0.981,\,0.987}$ & AUC \\
        jet mass       & $\bootci{0.983}{0.977,\,0.984}$ & $\bootci{0.953}{0.929,\,0.962}$ & $\bootci{0.993}{0.987,\,0.994}$ & $\bootci{0.990}{0.978,\,0.991}$ & $\bootci{0.981}{0.979,\,0.983}$ & $R^2$ \\
        jet mult       & $\bootci{0.923}{0.907,\,0.957}$ & $\bootci{0.827}{0.816,\,0.874}$ & $\bootci{0.971}{0.957,\,0.977}$ & $\bootci{0.962}{0.933,\,0.968}$ & $\bootci{0.921}{0.917,\,0.925}$ & $R^2$ \\
        jet $\tau_1$   & $\bootci{0.960}{0.949,\,0.967}$ & $\bootci{0.933}{0.916,\,0.942}$ & $\bootci{0.974}{0.964,\,0.977}$ & $\bootci{0.970}{0.950,\,0.976}$ & $\bootci{0.954}{0.947,\,0.957}$ & $R^2$ \\
        jet $\tau_2$   & $\bootci{0.875}{0.867,\,0.888}$ & $\bootci{0.855}{0.837,\,0.875}$ & $\bootci{0.924}{0.907,\,0.935}$ & $\bootci{0.927}{0.912,\,0.932}$ & $\bootci{0.900}{0.892,\,0.905}$ & $R^2$ \\
        jet $\tau_3$   & $\bootci{0.809}{0.764,\,0.837}$ & $\bootci{0.766}{0.714,\,0.779}$ & $\bootci{0.885}{0.864,\,0.894}$ & $\bootci{0.847}{0.806,\,0.863}$ & $\bootci{0.811}{0.793,\,0.839}$ & $R^2$ \\
        jet $\tau_{21}$& $\bootci{0.504}{0.447,\,0.557}$ & $\bootci{0.485}{0.436,\,0.517}$ & $\bootci{0.609}{0.557,\,0.637}$ & $\bootci{0.579}{0.557,\,0.600}$ & $\bootci{0.563}{0.536,\,0.583}$ & $R^2$ \\
        jet $\tau_{32}$& $\bootci{0.572}{0.549,\,0.605}$ & $\bootci{0.566}{0.545,\,0.587}$ & $\bootci{0.634}{0.603,\,0.653}$ & $\bootci{0.625}{0.609,\,0.641}$ & $\bootci{0.610}{0.591,\,0.630}$ & $R^2$ \\
        part $p_T$     & $\bootci{0.669}{0.657,\,0.696}$ & $\bootci{0.871}{0.853,\,0.875}$ & $\bootci{0.837}{0.808,\,0.852}$ & $\bootci{0.830}{0.824,\,0.845}$ & $\bootci{0.750}{0.709,\,0.757}$ & $R^2$ \\
        \textbf{part $\eta$} & $\bootci{-0.002}{-0.006,\,0.000}$ & $\bootci{0.000}{-0.005,\,0.005}$ & $\bootci{-0.006}{-0.011,\,-0.002}$ & $\bootci{0.151}{0.119,\,0.198}$ & $\bootci{0.287}{0.267,\,0.310}$ & $R^2$ \\
        part $\phi$    & $\bootci{-0.002}{-0.005,\,0.000}$ & $\bootci{-0.004}{-0.008,\,-0.001}$ & $\bootci{-0.004}{-0.008,\,-0.001}$ & $\bootci{-0.006}{-0.011,\,-0.002}$ & $\bootci{-0.005}{-0.008,\,-0.003}$ & $R^2$ \\
        part $E$       & $\bootci{0.557}{0.543,\,0.579}$ & $\bootci{0.739}{0.722,\,0.750}$ & $\bootci{0.722}{0.675,\,0.734}$ & $\bootci{0.712}{0.693,\,0.724}$ & $\bootci{0.639}{0.611,\,0.653}$ & $R^2$ \\
        part quartile  & $\bootci{0.763}{0.720,\,0.800}$ & $\bootci{0.723}{0.688,\,0.744}$ & $\bootci{0.760}{0.738,\,0.807}$ & $\bootci{0.738}{0.702,\,0.766}$ & $\bootci{0.803}{0.798,\,0.806}$ & Acc. \\
        part $\Delta R$& $\bootci{0.910}{0.901,\,0.922}$ & $\bootci{0.948}{0.936,\,0.956}$ & $\bootci{0.935}{0.928,\,0.946}$ & $\bootci{0.986}{0.981,\,0.986}$ & $\bootci{0.945}{0.926,\,0.947}$ & $R^2$ \\
        \bottomrule
      \end{tabular}
    }
  \end{center}
  \vskip -0.1in
\end{table}

\begin{figure}[h!]
  \begin{center}
    \includegraphics[width=\textwidth]{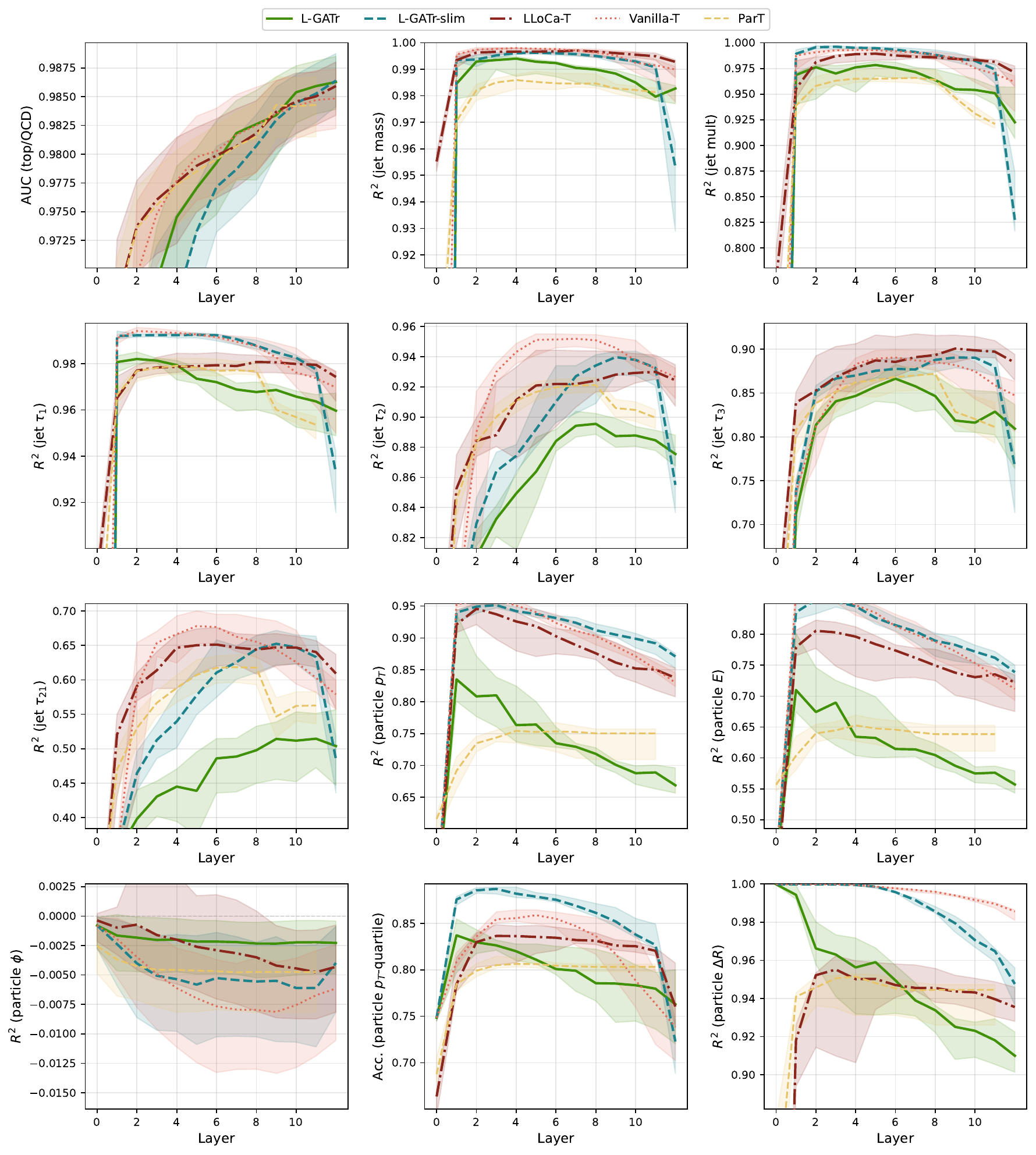}
    \caption{All remaining $\hs$ probe trajectories across all five TopTagging
      models ($4\times3$ grid; particle $\eta$ and jet $\tau_{32}$ shown in
      the main text are omitted).
      \textbf{Row 1}: top/QCD AUC, jet mass, jet multiplicity.
      \textbf{Row 2}: $\tau_1$, $\tau_2$, $\tau_3$.
      \textbf{Row 3}: $\tau_{21}$, particle $p_T$, particle $E$.
      \textbf{Row 4}: particle $\phi$, $p_T$-rank quartile, particle $\Delta R$.
      Shaded bands: combined 95\% bootstrap CIs pooled across 3 seeds for all models.
      Particle $\phi$ is near zero for all models, consistent with
      approximate azimuthal symmetry of the dataset.}
    \label{fig:hs_probe_extra}
  \end{center}
\end{figure}

Figure~\ref{fig:hs_probe_extra} collects the remaining $\hs$ probe trajectories
not shown in the main text, including top/QCD, jet mass, jet multiplicity,
$\tau_1$, $\tau_2$, $\tau_3$, $\tau_{21}$, particle $p_T$, particle $E$,
particle $\phi$, particle quartile and particle $\Delta R$.

\FloatBarrier
\section{Multivector Invariant Probes}
\label{app:mv_probes_full}

Tables~\ref{tab:mv_jet_full} and~\ref{tab:mv_particle_full} report linear probe
scores for all 832 Lorentz-invariant features at the final layer, broken down
by grade group: scalar-like (G0+G4), vector-like (G1+G3), bivector (G2), and
all 832 features combined.
Jet-level entries use combined-bootstrap medians with 95\% CIs pooled across 3
seeds (600 draws); $\hs$ uses seed mean $\pm$ std.
Particle-level entries use seed mean $\pm$ std throughout (3 seeds).

\textbf{Jet-level probes (Table~\ref{tab:mv_jet_full}).}
The ``All'' column (all 832 features combined) matches or exceeds $\hs$ for
regression targets; classification AUC is essentially identical (All
$= \bootci{0.985}{0.982,\,0.987}$ vs.\ $\hs = 0.987 \pm 0.000$).
For $\tau_{21}$, the full invariant set (All $= \bootci{0.648}{0.581,\,0.680}$)
notably exceeds $\hs$ ($0.494 \pm 0.040$), suggesting that $\tau_{21}$
information is encoded in the inter-grade structure of $\hmv$ but less
linearly accessible from $\hs$ alone.
The scalar-like and bivector groups have similar scores for classification
(G0$+$G4 $= \bootci{0.986}{0.984,\,0.989}$; from the combined bootstrap for
individual grades: G0 $= \bootci{0.986}{0.984,\,0.988}$,
G4 $= \bootci{0.986}{0.983,\,0.989}$), confirming that the model concentrates
discriminative invariants into the G0 and G4 blade values.

\begin{table}[H]
  \caption{Geometric-algebra-invariant probe scores at the final layer
    for \lgatr{} on TopTagging, broken down by grade group and target.
    Metric: $R^2$ for regression, AUC for classification.
    Multivector columns: combined-bootstrap median [95\% CI] (600 pooled
    draws across 3 seeds). $\hs$: seed mean $\pm$ std.}
  \label{tab:mv_jet_full}
  \vskip 0.1in
  \begin{center}
    \begin{small}
      \begin{tabular*}{\textwidth}{@{\extracolsep{\fill}}lccccc c}
        \toprule
        Target & $\hs$ & scalar-like & vector-like & bivector & All & Metric \\
        \midrule
        top/QCD        & $0.987{\pm}0.000$ & $\bootci{0.986}{0.984,\,0.989}$ & $\bootci{0.982}{0.979,\,0.985}$ & $\bootci{0.966}{0.961,\,0.971}$ & $\bootci{0.985}{0.982,\,0.987}$ & AUC \\
        jet mass       & $0.980{\pm}0.004$ & $\bootci{0.943}{0.933,\,0.957}$ & $\bootci{0.983}{0.955,\,0.989}$ & $\bootci{0.883}{0.858,\,0.923}$ & $\bootci{0.989}{0.980,\,0.993}$ & $R^2$ \\
        jet mult       & $0.913{\pm}0.017$ & $\bootci{0.905}{0.897,\,0.926}$ & $\bootci{0.950}{0.840,\,0.958}$ & $\bootci{0.802}{0.714,\,0.822}$ & $\bootci{0.974}{0.921,\,0.977}$ & $R^2$ \\
        jet $\tau_1$   & $0.958{\pm}0.006$ & $\bootci{0.924}{0.917,\,0.933}$ & $\bootci{0.958}{0.937,\,0.962}$ & $\bootci{0.878}{0.845,\,0.897}$ & $\bootci{0.971}{0.964,\,0.974}$ & $R^2$ \\
        jet $\tau_2$   & $0.872{\pm}0.007$ & $\bootci{0.832}{0.821,\,0.863}$ & $\bootci{0.914}{0.865,\,0.922}$ & $\bootci{0.761}{0.721,\,0.824}$ & $\bootci{0.925}{0.906,\,0.938}$ & $R^2$ \\
        jet $\tau_3$   & $0.802{\pm}0.022$ & $\bootci{0.773}{0.748,\,0.802}$ & $\bootci{0.861}{0.802,\,0.875}$ & $\bootci{0.668}{0.613,\,0.749}$ & $\bootci{0.893}{0.864,\,0.899}$ & $R^2$ \\
        jet $\tau_{21}$& $0.494{\pm}0.040$ & $\bootci{0.466}{0.434,\,0.488}$ & $\bootci{0.597}{0.510,\,0.626}$ & $\bootci{0.473}{0.417,\,0.501}$ & $\bootci{0.648}{0.581,\,0.680}$ & $R^2$ \\
        jet $\tau_{32}$& $0.558{\pm}0.010$ & $\bootci{0.587}{0.547,\,0.607}$ & $\bootci{0.616}{0.548,\,0.643}$ & $\bootci{0.493}{0.471,\,0.528}$ & $\bootci{0.658}{0.626,\,0.677}$ & $R^2$ \\
        \bottomrule
      \end{tabular*}
    \end{small}
  \end{center}
  \vskip -0.1in
\end{table}

\textbf{Particle-level probes (Table~\ref{tab:mv_particle_full}).}
The $\eta$ probe is near zero for all grade groups across all 3 seeds
(All: $-0.025 \pm 0.006$; $\hs$: $-0.003 \pm 0.002$), confirming that the
Lorentz-invariant features carry no substantial frame-dependent rapidity
information. The $\phi$ probe is similarly near zero for all groups
(All: $-0.021 \pm 0.002$).

\begin{table}[H]
  \caption{Particle-level geometric-algebra-invariant probe scores at the
    final layer for \lgatr{} on TopTagging. Format: seed mean $\pm$ std
    (3 seeds). The $\eta$ row is near zero for all grade groups (see
    Section~\ref{sec:mv_probes}).}
  \label{tab:mv_particle_full}
  \vskip 0.1in
  \begin{center}
    \begin{small}
      \begin{tabular*}{\textwidth}{@{\extracolsep{\fill}}lccccc c}
        \toprule
        Target & $\hs$ & scalar-like & vector-like & bivector & All & Metric \\
        \midrule
        part $p_T$           & $0.667{\pm}0.019$ & $0.544{\pm}0.049$ & $0.904{\pm}0.028$ & $0.481{\pm}0.064$ & $0.916{\pm}0.021$ & $R^2$ \\
        \textbf{part $\eta$} & $\mathbf{-0.003{\pm}0.002}$ & $\mathbf{-0.004{\pm}0.002}$ & $\mathbf{-0.020{\pm}0.006}$ & $\mathbf{-0.009{\pm}0.002}$ & $\mathbf{-0.025{\pm}0.006}$ & $R^2$ \\
        part $\phi$          & $-0.002{\pm}0.001$ & $-0.002{\pm}0.001$ & $-0.013{\pm}0.001$ & $-0.006{\pm}0.000$ & $-0.021{\pm}0.002$ & $R^2$ \\
        part $E$             & $0.588{\pm}0.013$ & $0.482{\pm}0.068$ & $0.849{\pm}0.021$ & $0.410{\pm}0.058$ & $0.866{\pm}0.016$ & $R^2$ \\
        part quartile        & $0.761{\pm}0.028$ & $0.660{\pm}0.027$ & $0.790{\pm}0.006$ & $0.653{\pm}0.030$ & $0.819{\pm}0.005$ & Acc. \\
        part $\Delta R$      & $0.912{\pm}0.008$ & $0.719{\pm}0.018$ & $0.864{\pm}0.011$ & $0.668{\pm}0.071$ & $0.926{\pm}0.002$ & $R^2$ \\
        \bottomrule
      \end{tabular*}
    \end{small}
  \end{center}
  \vskip -0.1in
\end{table}

\begin{figure}[h!]
  \begin{center}
    \includegraphics[width=\textwidth]{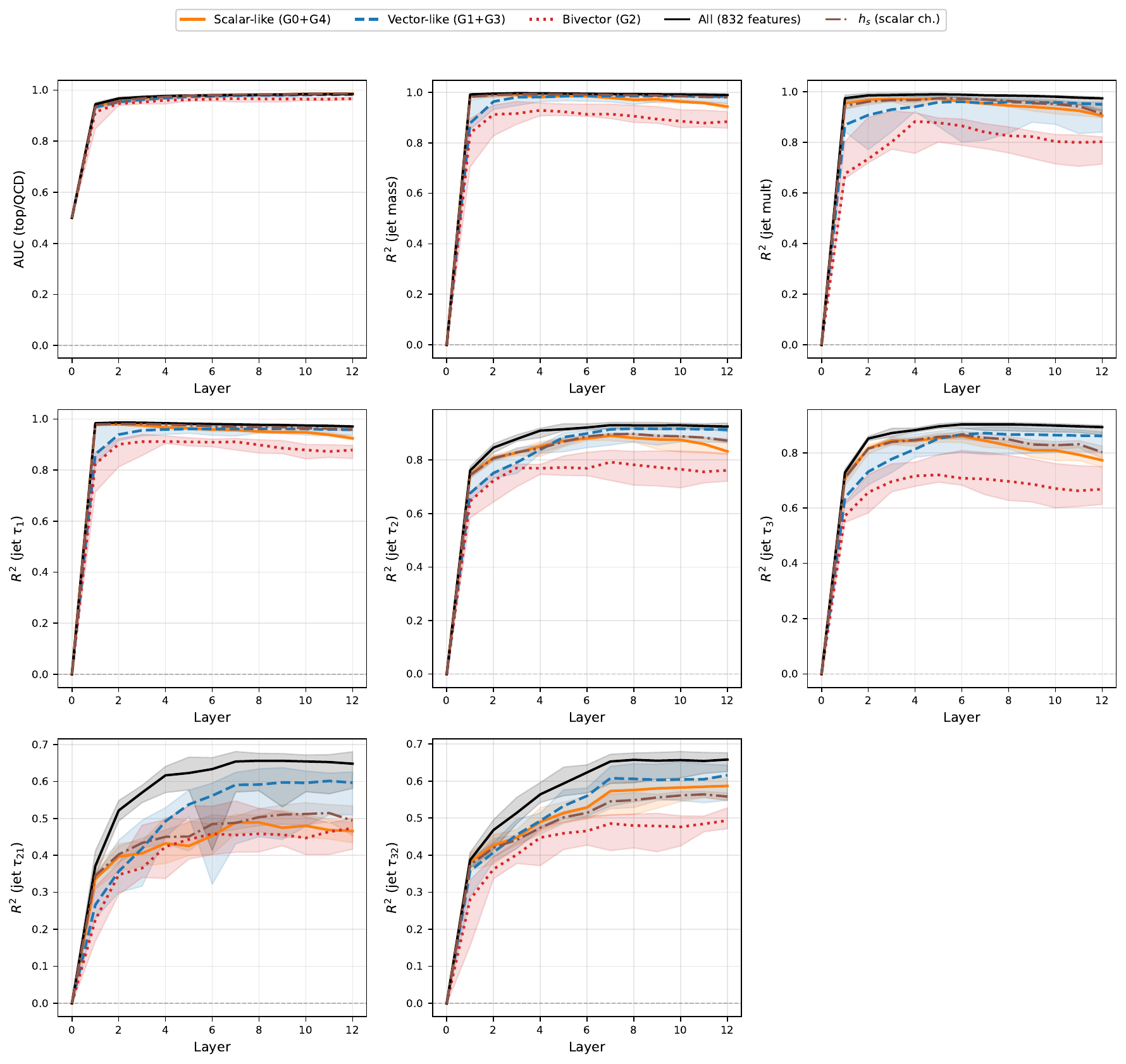}
    \caption{Per-layer geometric-algebra-invariant probe trajectories for all
      8 jet-level targets ($3\times3$ grid), broken down by grade group.
      Lines: scalar-like (G0+G4), vector-like (G1+G3), bivector (G2),
      all 832 features combined and $\hs$ scalar-channel reference.
      Shaded bands: combined 95\% bootstrap CIs for the multivector
      feature sets (600 pooled draws for jet-level targets); $\hs$ uses
      seed mean $\pm$ std.}
    \label{fig:mv_probes_jet}
  \end{center}
\end{figure}

\begin{figure}[h!]
  \begin{center}
    \includegraphics[width=\textwidth]{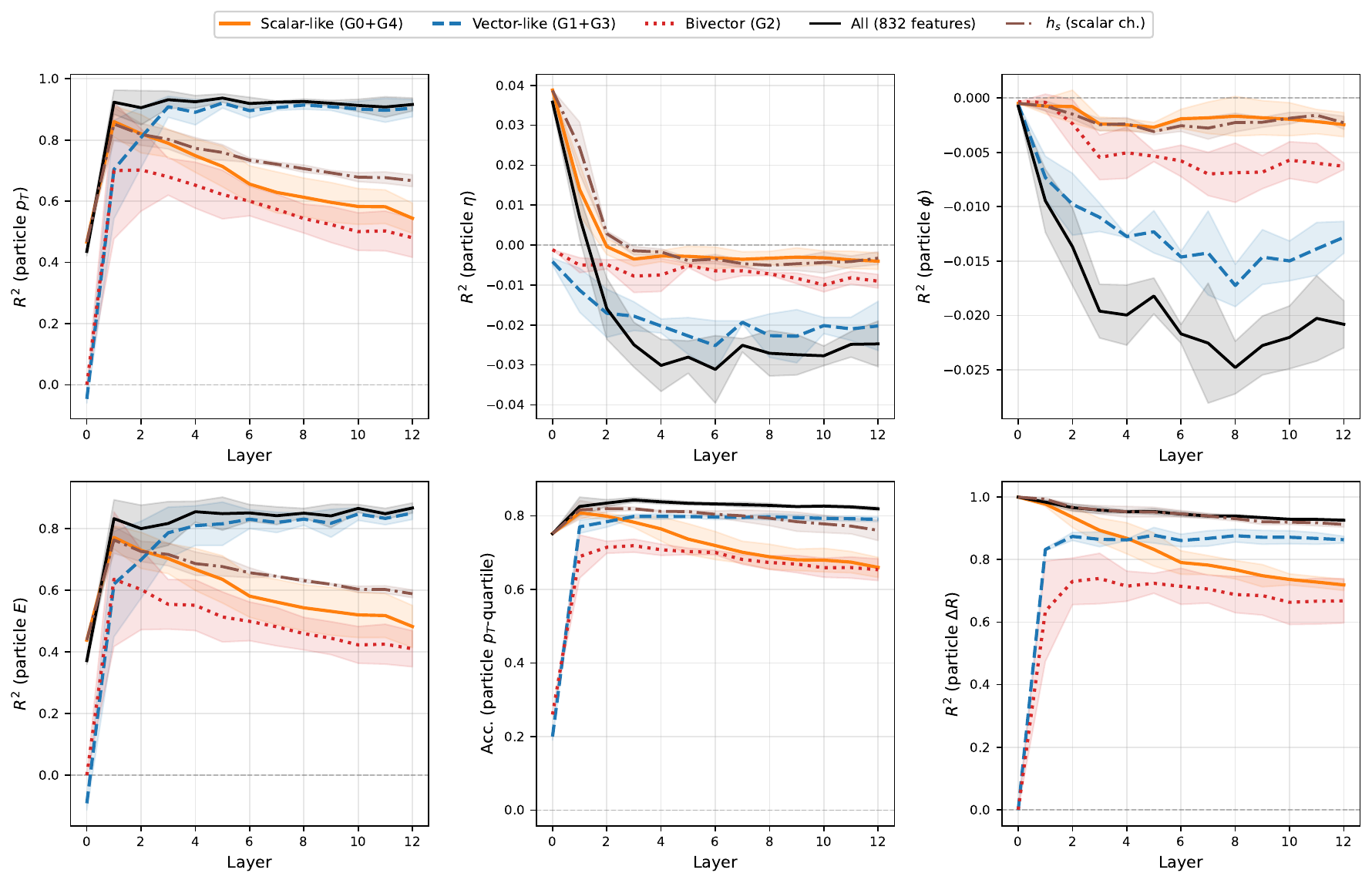}
    \caption{Per-layer geometric-algebra-invariant probe trajectories for all
      6 particle-level targets ($2\times3$ grid), broken down by grade group.
      Same line and color conventions as Figure~\ref{fig:mv_probes_jet}.
      Shaded bands: seed mean $\pm$ std (3 seeds).
      Particle $\eta$ and $\phi$ are near zero for all grade groups,
      confirming no frame-dependent information is encoded.}
    \label{fig:mv_probes_particle}
  \end{center}
\end{figure}

Figure~\ref{fig:mv_probes_particle} shows the corresponding particle-level
invariant-probe trajectories; as in the main text, particle $\eta$ and $\phi$
stay near zero across grade groups.

\FloatBarrier
\section{Grade Ablations}
\label{app:grade_ablations}

\textbf{Ablation methodology.}
Zero-grade ablation zeros the selected multivector grade or grade group at the outputs of all
13 network stages (the input linear layer plus 12 transformer blocks).
Keep-only ablation zeros all other grades or grade groups at those same hidden outputs.
Layer-resolved ablation zeros one grade or grade group at exactly one stage output. These
hooks operate on hidden multivectors rather than on the raw input embedding.
In all figures below, bars show combined 95\% bootstrap CIs pooled across
3 seeds (3\,000 draws); overlaid dots show per-seed $\Delta\mathrm{AUC}$; the right panel shows
seed mean $\pm$ std with per-seed thin lines for the dominant channel.

\subsection{Bivector=False Confirmation}
\label{app:biv_false}

\begin{figure}[h!]
  \begin{center}
    \includegraphics[width=\textwidth]{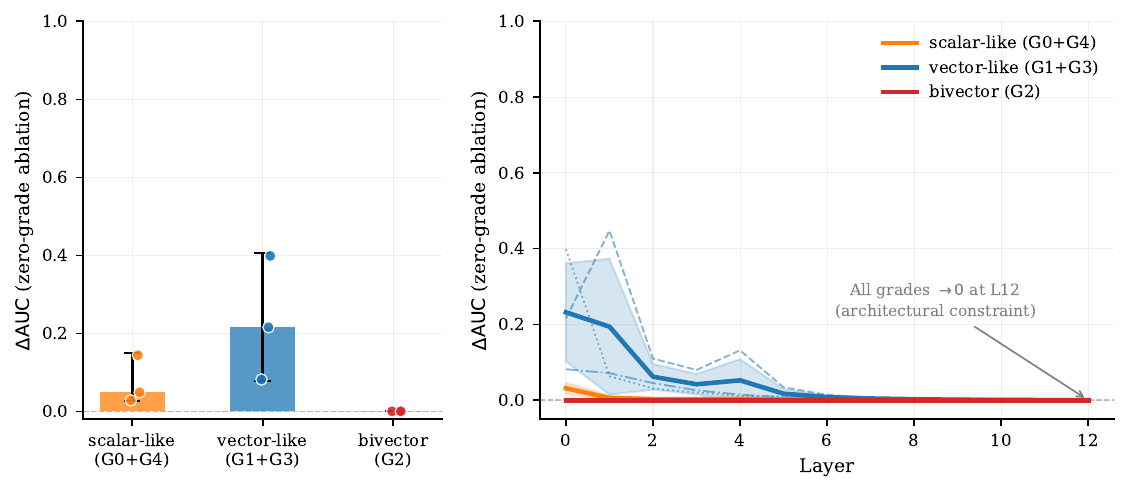}
    \caption{Grade ablations for \lgatr{} trained with bivectors architecturally
      disabled (3-group, G2 zeroed during training).
      \textbf{Left:} global zero-ablation $\Delta\mathrm{AUC}$ with 95\%
      combined bootstrap CIs (bars) and per-seed values (dots).
      \textbf{Right:} layer-resolved $\Delta\mathrm{AUC}$ (seed mean $\pm$
      std); thin lines show per-seed vector-like trajectories.
      Bivector bar is trivially zero; vector-like pattern mirrors the standard
      \lgatr{} seeds~1002/1003.}
    \label{fig:biv_false}
  \end{center}
\end{figure}

Figure~\ref{fig:biv_false} shows the grade ablation results for \lgatr{}
trained with bivectors architecturally disabled (G2 channel zeroed during training). The model achieves baseline AUC $0.9865 \pm 0.0002$,
indistinguishable from the standard \lgatr{} ($0.9867 \pm 0.0003$).
The bivector bar is trivially zero (G2 was never trained), while the
vector-like $\dauc$ per seed ($0.216$, $0.082$, $0.399$; mean $0.232 \pm 0.131$)
shows the same high cross-seed variance seen in the standard model.
The layer-resolved panel confirms that vector-like information again
concentrates at the early layers and decays with depth.

\subsection{\slim{} Channel Ablations}
\label{app:slim_grade}

\slim{} uses a different internal channel structure from \lgatr{}:
it retains only G0 scalars ($\hs$) and G1 4-vectors ($h_v$),
and its readout is the scalar channels at a global token.
Table~\ref{tab:slim_grade} reports channel ablations for \slim{} on TopTagging
(combined-bootstrap median [95\% CI]; baseline AUC
$= \bootci{0.986}{0.985,\,0.988}$).
Zeroing all scalar channels reduces \slim{}'s AUC to $0.500$ (chance,
$\dauc = \bootci{0.486}{0.485,\,0.488}$) because the readout is the global scalar token directly.
Zeroing the 4-vector channels gives a much larger effect than previously
reported ($\dauc = \bootci{0.585}{0.559,\,0.695}$), indicating that the vector channels
provide the geometric structure from which scalars are computed via the
Minkowski inner product in each block.

\begin{figure}[h!]
  \begin{center}
    \includegraphics[width=\textwidth]{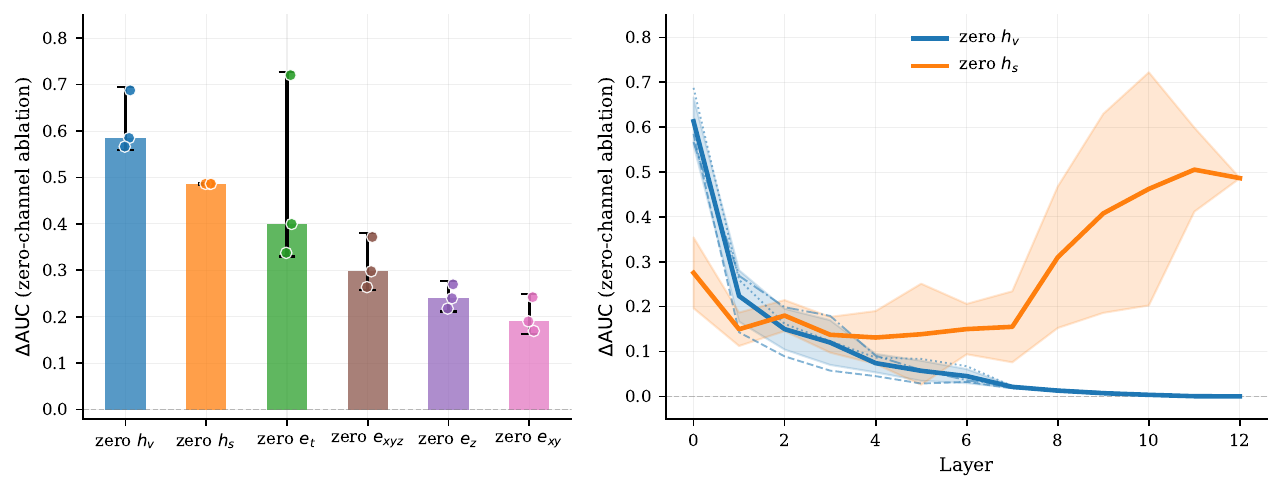}
    \caption{Channel ablations for \slim{} on TopTagging.
      \textbf{Left:} global $\Delta\mathrm{AUC}$ for all six interventions
      with combined 95\% bootstrap CIs (bars) and per-seed dots.
      Interventions ordered by decreasing impact: $h_v$ (all vectors),
      $\hs$ (all scalars), $e_t$ (energy), $e_{xyz}$ (3-momentum),
      $e_z$ (beam axis), $e_{xy}$ (transverse).
      \textbf{Right:} layer-resolved $\Delta\mathrm{AUC}$ for $\hs$ and $h_v$
      only (component ablations were not run per-layer); thin lines show
      per-seed $h_v$ trajectories.}
    \label{fig:slim_grade}
  \end{center}
\end{figure}

The component-level ablations reveal which 4-vector components matter most:
the energy component $e_t$ is most critical
($\dauc = \bootci{0.400}{0.330,\,0.727}$, with a wide seed band), followed by the
full 3-momentum $e_{xyz}$ ($\dauc = \bootci{0.299}{0.257,\,0.381}$). The beam-axis $e_z$ alone
($\dauc = \bootci{0.240}{0.211,\,0.277}$) is more important than transverse components $e_{xy}$
($\dauc = \bootci{0.190}{0.164,\,0.249}$), consistent with longitudinal momentum being more
discriminative for TopTagging.
Figure~\ref{fig:slim_grade} visualizes the same interventions globally and
layer-by-layer.

\begin{table}[H]
  \caption{Channel ablations for \slim{} on TopTagging. Format:
    combined-bootstrap median [95\% CI]; baseline AUC
    $= \bootci{0.986}{0.985,\,0.988}$.}
  \label{tab:slim_grade}
  \vskip 0.1in
  \begin{center}
    \begin{small}
            \begin{sc}
        \begin{tabular}{lcc}
          \toprule
          Channel zeroed & AUC (mean) & $\dauc$ \\
          \midrule
          \textbf{$\hs$ (all scalars)} & \textbf{0.500} & $\bootci{0.486}{0.485,\,0.488}$ \\
          $h_v$ (all vectors)          & 0.401          & $\bootci{0.585}{0.559,\,0.695}$ \\
          \midrule
          $e_t$ (energy component)     & 0.586          & $\bootci{0.400}{0.330,\,0.727}$ \\
          $e_z$ (beam axis)            & 0.746          & $\bootci{0.240}{0.211,\,0.277}$ \\
          $e_x, e_y$ (transverse)      & 0.796          & $\bootci{0.190}{0.164,\,0.249}$ \\
          $e_x, e_y, e_z$ (3-momentum) & 0.687          & $\bootci{0.299}{0.257,\,0.381}$ \\
          \bottomrule
        \end{tabular}
      \end{sc}
    \end{small}
  \end{center}
  \vskip -0.1in
\end{table}

\subsection{Subgroup=False Side Study}
\label{app:subgroup_false}

\begin{figure}[h!]
  \begin{center}
    \includegraphics[width=\textwidth]{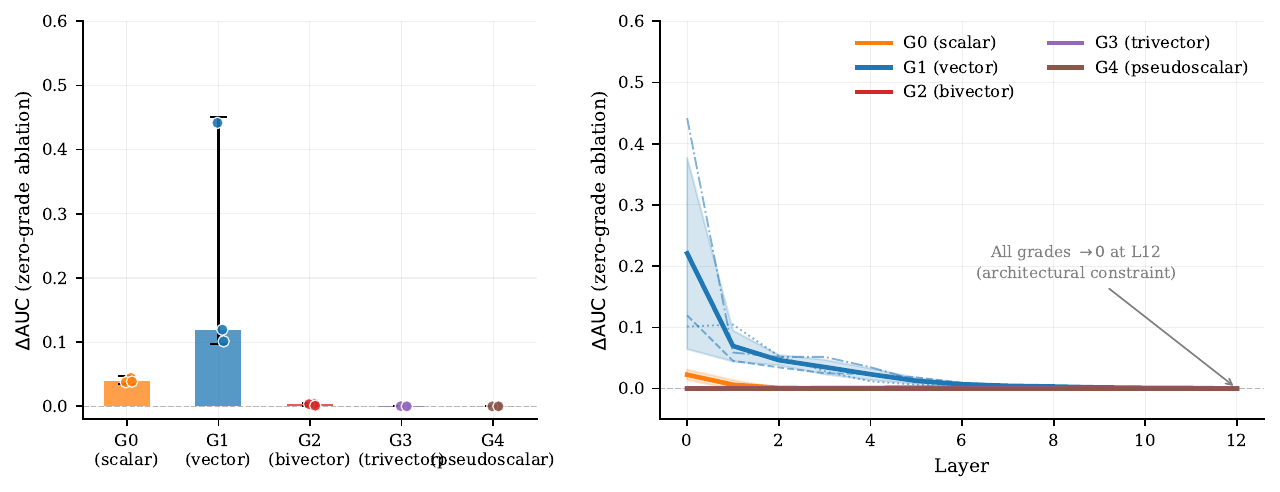}
    \caption{Grade ablations for \lgatr{} with the alternative subgroup setting
      (5 independent grades G0--G4).
      \textbf{Left:} global zero-ablation $\Delta\mathrm{AUC}$ with combined
      95\% bootstrap CIs and per-seed dots.
      \textbf{Right:} layer-resolved $\Delta\mathrm{AUC}$ (seed mean $\pm$ std);
      thin lines show per-seed G1 trajectories.
      G2, G3, and G4 are all negligible; G1 is again dominant with high
      cross-seed variance, consistent with the 3-group main result.}
    \label{fig:subgroup_false}
  \end{center}
\end{figure}

Table~\ref{tab:subgroup_false} and Figure~\ref{fig:subgroup_false} report
zero-grade ablations for \lgatr{} trained with the alternative subgroup
setting, which uses five independent grades rather than the three mixed
groups used in the main text. This side study confirms that the G2 $\approx 0$
finding is not an artifact of the subgroup mixing in the default setting.

G2 bivector $\dauc = \bootci{0.003}{0.001,\,0.005}$, essentially zero,
confirming the main-text 3-group result.
G1 vector is again dominant with high variance
($\dauc = \bootci{0.120}{0.097,\,0.451}$),
consistent with the multiple-pathway interpretation.
G3 trivector and G4 pseudoscalar are both negligible ($\dauc < 0.001$),
indicating parity-odd components carry no load-bearing information for
TopTagging.

\begin{table}[H]
  \caption{Zero-grade ablations for \lgatr{} on TopTagging with the alternative
    subgroup setting (5 independent grades).
    Format: combined-bootstrap median [95\% CI]; per-seed values in brackets.
    Baseline AUC: $\bootci{0.986}{0.984,\,0.988}$.}
  \label{tab:subgroup_false}
  \vskip 0.1in
  \begin{center}
    \begin{small}
            \begin{sc}
        \begin{tabular}{lc}
          \toprule
          Grade & Zero $\dauc$ \\
          \midrule
          G0 scalar         & $\bootci{0.039}{0.035,\,0.047}$\ [$0.044, 0.038, 0.039$] \\
          G1 vector         & $\bootci{0.120}{0.097,\,0.451}$\ [$0.119, 0.442, 0.101$] \\
          \textbf{G2 bivector}       & $\bootci{0.003}{0.001,\,0.005}$\ [$0.004, 0.003, 0.001$] \\
          G3 trivector      & $\bootci{0.000}{0.000,\,0.001}$         \\
          G4 pseudoscalar   & $\bootci{0.000}{0.000,\,0.000}$         \\
          \bottomrule
        \end{tabular}
      \end{sc}
    \end{small}
  \end{center}
  \vskip -0.1in
\end{table}

\FloatBarrier
\section{Equivariance Details}
\label{app:equivariance_bootstrap}

\textbf{Equivariance uncertainty.}
For the random-transform equivariance summary, each model is evaluated on
200 test jets with 5 random Lorentz transforms per seed. The tabulated
intervals pool 1,000 bootstrap draws from each of the 3 independently trained
seeds, for 3,000 pooled draws per model. Table~\ref{tab:equivariance_full}
reports the full uncertainty summary for the equivariance experiment in Section~\ref{sec:equivariance}.

\begin{table}[H]
  \caption{Mean relative logit change under random Lorentz transforms (200 jets
    $\times$ 5 random transforms per seed). Values are combined-bootstrap medians with 95\% CIs; seed ranges are shown
    separately. Lower is more equivariant.
    $\dagger$\lloca{} is equivariant via canonicalization; larger measured errors reflect numerical sensitivity of the canonicalization step.}
  \label{tab:equivariance_full}
  \vskip 0.1in
  \begin{center}
    \begin{small}
      \begin{sc}
        \begin{tabular}{lccc}
          \toprule
          Model & Bootstrap median [95\% CI] & Seed range & vs.\ \lgatr{} \\
          \midrule
          \lgatr{}          & $\bootci{5.2}{3.0,\,8.7}\times10^{-3}$ & $[0.31,\,12]\times10^{-3}$ & $1\times$ \\
          \slim{}           & $\bootci{3.9}{2.6,\,6.3}\times10^{-3}$ & $[2.9,\,4.9]\times10^{-3}$ & $0.75\times$ \\
          \lloca{}$^\dagger$ & $\bootci{0.131}{0.060,\,0.251}$        & $[0.071,\,0.26]$           & ${\sim}25\times$ \\
          \vanilla{}        & $\bootci{1.20}{0.98,\,1.63}$           & $[1.03,\,1.56]$            & ${\sim}230\times$ \\
          \part{}           & $\bootci{3.34}{2.57,\,4.29}$           & $[2.04,\,4.79]$            & ${\sim}650\times$ \\
          \bottomrule
        \end{tabular}
      \end{sc}
    \end{small}
  \end{center}
  \vskip -0.1in
\end{table}

\subsection{Per-Layer Equivariance}

Table~\ref{tab:equivariance_layers} reports the per-layer equivariance
errors for \lgatr{} on TopTagging (200 jets, 5 random Lorentz transforms each).

At layer 0 (the input projection), $\hs$ invariance is exact to machine
precision ($0$) and $\hmv$ equivariance error is $5 \times 10^{-8}$
(float32 machine epsilon), so the input projection is exactly equivariant.
Across layers 1--12, errors accumulate to $10^{-4}$--$5\times 10^{-3}$
for $\hs$ and $7\times 10^{-3}$--$10^{-2}$ for $\hmv$, consistent with
floating-point drift across 12 GATrBlock operations.
The na\"{i}ve baseline (treating a boosted input as unchanged) gives errors
of $\sim 1.3$ throughout, confirming that the equivariant architecture
reduces the effective error by approximately $100\times$ at all layers.

We cap the main-text sweep at $\gamma \leq 5$ because stronger boosts are
poorly supported by the TopTagging rapidity range and are more sensitive to
implementation-level numerical effects.

\begin{table}[H]
  \caption{Per-layer equivariance errors for \lgatr{} on TopTagging
    (200 jets $\times$ 5 random Lorentz transforms).
    $\hs$ error: relative change in scalar channels (should be 0).
    $\hmv$ equivariance: relative change after applying the correct
    group action. Na\"{i}ve: error if no transformation applied.}
  \label{tab:equivariance_layers}
  \vskip 0.1in
  \begin{center}
    \begin{small}
      \begin{tabular}{lccc}
        \toprule
        Layer & $\hs$ error & $\hmv$ equiv. & Na\"{i}ve \\
        \midrule
        L0 (input projection) & $0$ (exact) & $5\times10^{-8}$ & $1.36$ \\
        L1--L12                  & $10^{-4}$--$5\times10^{-3}$ & $7\times10^{-3}$--$10^{-2}$ & $\sim 1.3$ \\
        \midrule
        Output logit & \multicolumn{2}{c}{$5.17\times10^{-3}$ (0.52\%)} & \\
        \bottomrule
      \end{tabular}
    \end{small}
  \end{center}
  \vskip -0.1in
\end{table}

Table~\ref{tab:boost_sweep} reports the boost sweep for \lgatr{}.

\begin{table}[H]
  \caption{Boost sweep for \lgatr{} on TopTagging (pure boosts in a random
    direction, 50 jets per seed). Values are combined-bootstrap medians with
    95\% CIs pooled across 3 seeds. $\hmv$ error: relative change in multivector
    hidden states. Logit error: relative change in output logit.}
  \label{tab:boost_sweep}
  \vskip 0.1in
  \begin{center}
    \begin{small}
      \begin{sc}
        \begin{tabular}{lcc}
          \toprule
          $\gamma$ & $\hmv$ error & Logit error \\
          \midrule
          1.0  & $\bootci{0}{0,\,0}$ & $\bootci{0}{0,\,0}$ \\
          1.25 & $\bootci{3.5}{2.7,\,4.5}\times10^{-3}$ & $\bootci{1.9}{1.3,\,2.6}\times10^{-3}$ \\
          1.5  & $\bootci{5.8}{4.4,\,7.5}\times10^{-3}$ & $\bootci{3.3}{2.1,\,4.8}\times10^{-3}$ \\
          1.75 & $\bootci{7.4}{5.8,\,9.4}\times10^{-3}$ & $\bootci{3.8}{2.5,\,5.5}\times10^{-3}$ \\
          2.0  & $\bootci{8.9}{7.0,\,12}\times10^{-3}$ & $\bootci{4.8}{3.2,\,6.9}\times10^{-3}$ \\
          2.5  & $\bootci{1.4}{1.1,\,1.7}\times10^{-2}$ & $\bootci{8.9}{5.1,\,15}\times10^{-3}$ \\
          3.0  & $\bootci{2.1}{1.7,\,2.6}\times10^{-2}$ & $\bootci{1.2}{0.77,\,1.7}\times10^{-2}$ \\
          4.0  & $\bootci{3.1}{2.5,\,3.7}\times10^{-2}$ & $\bootci{1.8}{1.1,\,2.9}\times10^{-2}$ \\
          5.0  & $\bootci{4.7}{3.9,\,5.6}\times10^{-2}$ & $\bootci{3.0}{1.9,\,4.3}\times10^{-2}$ \\
          \bottomrule
        \end{tabular}
      \end{sc}
    \end{small}
  \end{center}
  \vskip -0.1in
\end{table}

\end{document}